\documentclass[journal]{IEEEtran}
\usepackage{cite}
\usepackage{times} 
\usepackage{helvet} 
\usepackage{courier} 
\usepackage[hyphens]{url} 
\usepackage{graphicx} 
\usepackage{algorithm}
\usepackage{indentfirst}
\usepackage{algpseudocode}
\usepackage{booktabs}
\usepackage{multirow}
\usepackage{changepage}
\usepackage{setspace}
\usepackage{subfig}
\usepackage{cite}
\usepackage{amsmath}
\usepackage{amsmath}
\newtheorem{theorem}{Theorem}

\newtheorem{definition}{Definition}
\newtheorem{corollary}{Corollary}
\usepackage{multirow}
\usepackage{amssymb}

\usepackage{listings}
\usepackage{color}
\usepackage{color}
\usepackage{tabularx}

\usepackage{eqparbox}

\ifCLASSINFOpdf
\else
\fi
\begin{document}
%
\title{Self-supervised Cross-silo Federated Neural Architecture Search}
%
%
%

\author{Xinle~Liang,~\IEEEmembership{Member,~IEEE}, Yang~Liu\thanks{\*},~\IEEEmembership{Member,~IEEE},
Jiahuan~Luo, Yuanqin~He, Tianjian Chen, 
and~Qiang~Yang,~\IEEEmembership{Fellow,~IEEE}
\IEEEcompsocitemizethanks{\IEEEcompsocthanksitem Xinle Liang and Yang Liu are co-first authors. Yang Liu and Qiang Yang are the corresponding authors. Email: yangliu@webank.com, qyang@cse.ust.hk}
\IEEEcompsocitemizethanks{\IEEEcompsocthanksitem Xinle Liang, Yang Liu, Jiahuan Luo, Yuanqin He, Tianjian Chen and Qiang Yang are with Department of Artificial Intelligence, Webank, Shenzhen, China. Qiang Yang is also affliated with Hong Kong University of Science and Technology.}
}

%
%

\markboth{}%
{Shell \MakeLowercase{\textit{et al.}}: Bare Demo of IEEEtran.cls for Computer Society Journals}
%



\maketitle

\begin{abstract}
Federated Learning (FL) provides both model performance and data privacy for machine learning tasks where samples or features are distributed among different parties. In the training process of FL, no party has a global view of data distributions or model architectures of other parties. Thus the manually-designed architectures may not be optimal. In the past, Neural Architecture Search (NAS) has been applied to FL to address this critical issue. However, existing Federated NAS approaches require prohibitive communication and computation effort, as well as the availability of high-quality labels. In this work, we present Self-supervised Vertical Federated Neural Architecture Search (SS-VFNAS) for automating FL where participants hold feature-partitioned data, a common cross-silo scenario called Vertical Federated Learning (VFL). In the proposed framework,  each party first conducts NAS using self-supervised approach to find a local optimal architecture with its own data. Then, parties collaboratively improve the local optimal architecture in a VFL framework with supervision.  We demonstrate experimentally that our approach has superior performance, communication efficiency and privacy compared to Federated NAS and is capable of generating high-performance and highly-transferable heterogeneous architectures even with insufficient overlapping samples, providing automation for those parties without deep learning expertise.\footnote{This work has been submitted to the IEEE for possible publication. Copyright may be transferred without notice, after which this version may no longer be accessible.}
\end{abstract}


%
\IEEEpeerreviewmaketitle

\section{Introduction}
\IEEEPARstart{D}{ata} privacy has become one of the main research topics in machine learning.  In some commercial scenarios, data-sharing, such as the sharing of patients' medical data or the sharing of consumers' financial data, may bring prohibitive economical costs or legal risks.  Privacy-preserving machine learning (PPML) is devoted to building high-performance models without the leakage of data, data structures, or even model structures.

McMahan et al. \cite{mcmahan2017communication} proposed Federated Learning (FL) to train local language models on millions of mobile devices without collecting user's private data. Extending the cross-device FL concept, the cross-silo Federated Learning setting deals with collaborative machine learning with privacy preservation among different organizations \cite{yang2019federated,kairouz2019federated}. 
Over the past, the research and industrial community has enabled FL with open-sourced modeling tools, including TensorFlow Federated \cite{tensorflowfederated}, PySyft \cite{ryffel2018generic}, PaddleFL \cite{paddlefl} and FATE \cite{webankfate} etc. Yang et al. \cite{yang2019federated}  presented a comprehensive survey on different FL scenarios, and categorized them into the following:
\begin{enumerate}
\item \textit{Horizontal Federated Learning (HFL):} HFL describes the scenarios where different parties having data of the same features collaboratively train a global model. \cite{mcmahan2017communication,rothchild2020fetchsgd};
\item \textit{Vertical Federated Learning (VFL):} VFL describes the scenarios where different organizations with a common set of users train a cooperative model to better utilize the data with distributed features \cite{yang2019federated,Kairouz2019AdvancesAO,Vepakomma2018Split,gupta2018distributed,Hu2019LearningPO};
\item \textit{Federated Transfer Learning (FTL):} FTL applies to the cases where the data sets are simultaneously different in samples and features but share some common knowledge \cite{Chen2020FedHealthAF,Liu2020ASF};
\end{enumerate}

Our work is targeted at improving the model performance and communication efficiency in VFL, and specifically the multi-domain image classification application, where model designers are required to build high-performance networks on multi-source image data where data privacy needs to be protected. 
One typical example comes from medical applications  where in order to investigate the nature of diseases, such as Alzheimer’s Disease(AD), multiple modalities of the diagnosis data, including Magnetic Resonance Imaging (MRI) and Positron Emission Tomography (PET) are used together to improve the performance of deep learning models\cite{ebrahimighahnavieh2020deep}. 

\begin{figure}[htb]
\center{\includegraphics[width=8cm] {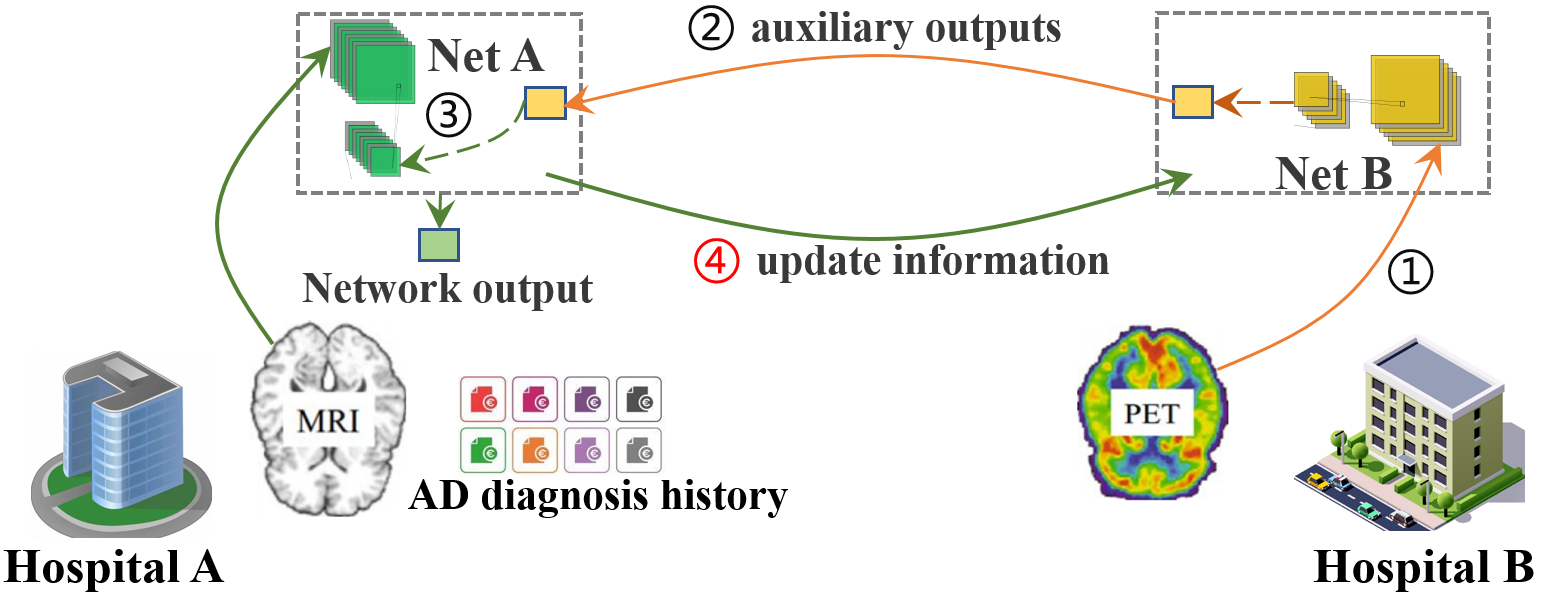}}
\caption{ Cross-hospital AD diagnosis VFL system. Net A and B are separately maintained by different hospitals to extract complementary information of PET and MRI data.}
\label{example}
\end{figure}

However, in real-life medical systems, each hospital may have only MRI or PET data and these approaches may not be practical\cite{Changqing2018Multi}. 
In order to train a more accurate and robust model, these hospitals may seek to cooperate with other hospitals without violating patient's privacy in a VFL framework, see Figure \ref{example}. 

Due to the complex nature of VFL, system designers may encounter lots of obstacles when deploying VFL in real-life systems. Our work is motivated by the following challenges observed:
\begin{itemize}

\item \textit{Uninspectable data and lack of expertise:} Usually, some VFL participants lack the professional skills of designing neural network structures. Moreover, data are often uninspectable and private in VFL. Therefore without a global view of the data structure, it is challenging to make design choices that are optimal even for modeling experts\cite{kairouz2019federated};
\item \textit{Incapability of adjusting to resource constrained edge devices:} Usually, the manual architectures rely heavily on researchers' knowledge and are often too complex to be adaptively deployed onto different types of resource constrained devices;
\item \textit{Unacceptable communication complexity:} Training VFL algorithms often require expensive communication overhead  \cite{Liu2019ACE};
\item \textit{Data leakage risks:} In the training process, the exchange of messages among parties may leak information on raw data \cite{wu2020mitigatingba,wei2020aff};
\end{itemize}

Our solution for tackling the challenges above relies on  Neural Architecture Search (NAS), which has become a practical solution for automating deep learning process for  system designers without strong prior knowledge \cite{elsken2019neural}. This solution falls into the realm of conducting NAS tasks under VFL conditions and the key challenge is how to efficiently and simultaneously gain the optimal model architectures for all parties without exchanging raw data or local model information. We term this \textit{Vertical Federated Neural Architecture Search} (VFNAS). We will demonstrate the details of this approach in the following sections. 
In the past, NAS has been applied to HFL scenarios with privacy considerations \cite{zhu2020real-time,singh2020differentially,xu2020neural,chaoyang2020fednas}. However such approaches require heavy communication of architecture and weight parameters. 

Similarly, a straightforward integration of NAS and VFL can be unsatisfactory. First, a few studies \cite{Liu2019ACE,Lu2020CommunicationefficientFL,Asad2020EvaluatingTC} have shown that, when processing complex data (for example, the high-resolution medical images), the communication efficiency of a Federated NAS framework can be prohibitively low. Compared to a conventional FL framework which only communicates model weight parameters, the training process of gradient-based NAS tasks require the optimization and communication of both weight and architecture parameters on the train and validation dataset, respectively, adding further burdens to communication overhead. 
Secondly, it has been shown that in the HFL scenarios where gradients are transmitted and exposed instead of raw data, adversarial parties may recover essential information from exchanged gradients \cite{zhu2019deeplf}. In VFL scenarios, since each party has its own data and sub-model, only the final layer of a local neural architecture after forward propagation is exchanged. Although it has been shown that it is unlikely to recover exact raw data from such intermediate results \cite{Liu2019ACE,gupta2018distributed}, there is room for further privacy enhancement. Thirdly, the naive VFNAS approach requires availability of sufficient overlapping samples and high quality labels, and is difficult to generalize to other tasks.

To improve the communication efficiency and transferablity of VFNAS, we introduced self-supervised NAS, which is expected to generate promising weight and architecture parameters without communications among VFL parties and thus help the downstream VFNAS converge faster. This approach is named \textit{Self-Supervised Vertical Federated Neural Architecture Search (SS-VFNAS)}. At last, we apply  Differential Privacy (DP) and perform analysis on the impact of privacy and utility.

The contribution of the work can be summarized as follows:
\begin{enumerate}
\item We present VFNAS, a generalized formulation of NAS tasks in the VFL setting which enables VFL parties to simultaneously optimize heterogeneous network architectures with uninspectable data, various model complexity constraints, and with privacy preserved via differential privacy. 
\item In order to improve communication and privacy of VFNAS, we further present SS-VFNAS, where each party independently performs self-supervised pre-searching of both weight and architecture parameters, and then jointly perform supervised VFNAS as downstream task. We show that self-supervised training can enable parties to find a reasonable architecture as well as initial weights, whereas collaboratively fine-tuning of both architecture and weight parameters are necessary to achieve optimal supervised performance. 
\item We propose two benchmark image datasets for VFL frameworks and demonstrate experimentally that this approach outperforms several baselines including self-supervised local NAS training (SSNAS), naive VFNAS and end-to-end co-training of self-supervised representations and supervised tasks (SSNAS\_E2E), and the advantage of SS-VFNAS is more pronounced when learning with limited number of overlapping samples. Overall, SS-VFNAS achieves superior communication efficiency, privacy and transferability than the native federated NAS approach (VFNAS).  
\end{enumerate}

This work is organized as follows: Section \ref{sec:relatedworks} is a review of related works on VFL, Federated NAS and Self-supervised NAS techniques. The formulation of VFNAS is presented in Section \ref{sec:formulation}. In Section \ref{sec:methods}, we describe the VFNAS algorithms, SS-VFNAS and differential privacy approaches. Then, in Section \ref{sec:experiments}, we demonstrate the essentiality and performance of the proposed approach. We summarize the conclusions and  future research directions in Section \ref{sec:future}.

\section{Related Works}\label{sec:relatedworks}

\subsection{Vertical Federated Learning}\label{subsec:fl}

Federated Learning (FL), first introduced by \cite{mcmahan2017communication}, is a distributed machine learning setting focusing on data locality and privacy.  Over the past, FL has received significant interest from both research and industrial communities. The original FL framework \cite{mcmahan2017communication} optimize a consensus machine learning model based on data with the same features across millions of users/devices. Google has applied FL to its Gboard mobile keyboard applications \cite{Hard2018federated,Ramaswamy2019federated,Chen2019federated}.  

Cross-silo Federated Learning considers data from multiple organizations. Vertical FL, especially, considers the scenario where multiple parties having the same set of the users but different feature attributes. Liu et al. \cite{Liu2019ACE} studied the communication efficiency in VFL, and presented a Federated Stochastic Block Gradient Descent (FedBCD), which enables VFL participant to execute multiple local updates with the proximal term. 
Liu et al. \cite{Liu2020AsymmetricalVF} studied the asymmetrical vertical federated learning problem, where different participants have non-identical privacy concerns over the samples ID. 
Feng and Yu \cite{Feng2020MultiParticipantMV} investigated the Multi-participants Multi-class VFL (MMVFL) problem, which enables the label-sharing operations among different participants in privacy-preserving manners. Yang et al. \cite{Yang2019ParallelDL} studied the VFL problem without the presence of third-party coordinator, and presented a solution for parallel distributed logistic regression. 

Split Neural Network (SplitNN) \cite{Vepakomma2018Split,gupta2018distributed,vepakomma2019reducing,poirot2019split,Ceballos2020SplitNNdrivenVP,Abuadbba2020CanWU} is another framework that can be employed for dealing with VFL problems. 
In this framework, multiple network partitions interactively exchange network intermediate outputs and their corresponding gradients in a forward-and-backward manner, such that all the distributed local network models are separately updated. 



\subsection{Federated NAS}\label{subsec:fednas}

In recent years, NAS approaches have been proven to be a powerful autonomous tool to facilitate the process of designing complex deep learning networks with no predefined model architectures \cite{liu2018progressive,zoph2018learning,real2018regularized}. In literature, there are basically three kinds of architecture searching methods: evolutionary algorithms based \cite{zoph2018learning}, reinforcement-learning-based \cite{liu2018progressive} and gradient-based \cite{liu2018darts:}. RL-based or evolutionary algorithm based NAS techniques may consume more than 1000 GPU days to achieve the state-of-the-art results \cite{liu2018progressive,Liang2019EvolutionaryNA}. Therefore, for the sake of computation efficiency, we choose gradient-based NAS approaches, which are demonstrated to be capable of reducing the training and evaluating time to only a few hours (or days) or even running on mobile devices \cite{cai2020once,Cai2019ProxylessNASDN}.

Kairouz and McMahan \cite{kairouz2019federated} presented 
the practical necessity and promise to investigate NAS tasks under FL paradigm. Federated NAS is a recently emerging technique composition which utilizes the ability of NAS to automate designing optimal network structure, and the ability of FL framework to collaboratively train models with user data privacy preserved. With Federated NAS framework, multiple parties can collaboratively search for an optimal network architecture that yields the best performance on the validation dataset. This may greatly release the burdens of manually designing network structures separately.

Zhu and Jin \cite{zhu2020real-time} applied NAGA-II \cite{Deb2002fast} to the multi-objective problem in federated NAS tasks for simultaneously optimizing the model performance and local client's payload. He and Annavaram \cite{chaoyang2020fednas} invested FedAvg \cite{mcmahan2017communication} algorithm on the federated NAS systems, where multiple local clients cooperatively search for an optimal model without sharing the local data. Similarly, Singh et al. \cite{singh2020differentially} studied DP-FNAS algorithm which aggregates the gradients from the local network, and uses differential privacy to protect the communication contents. 
Xu et al. \cite{xu2020neural} presented FedNAS which introduced NAS into the federated learning system, with the considerations of several key optimizations, including communication costs and local computation costs. 

All the  Federated NAS researches above fall into the categorization of empowering NAS approaches with FedAvg-like algorithm. The executions of these approaches are based on the assumption that all participants have a consensus that model architecture are the same and their parameters can be federated. This restricts the existing Federated NAS approaches to the HFL framework. Our work investigates NAS tasks within VFL scenario, which is capable of generating multiple heterogeneous networks with no consensus architecture from different participants.

\subsection{Self-Supervised NAS}\label{subsec:selfnas}

Most of the existing NAS researches focus on image classification tasks, such as CIFAR-10 \cite{krizhevsky2009learning} and ImageNet \cite{russakovsky2015imagenetls}. Recently, a few studies have shown that self-supervised NAS approaches can identify sufficiently-good network architectures without data annotations. Liu et al. \cite{Liu2020AreLN} 
proposed Self-Supervised NAS approach with various unsupervised objectives. 
Kaplan and Giryes \cite{Kaplan2020SelfsupervisedNA} further presented a contrastive self-supervised learning NAS architecture. 
Based on these researches, we hypothesize that self-supervised NAS can be used in VFNAS as a pretraining or co-training part of the VFNAS process to improve the overall communication and privacy of VFL.

\section{Problem Formulation}\label{sec:formulation}

In this section, we first introduce the VFL framework adopted in \cite{Liu2019ACE} where each party holds an unique set of features of common users. Suppose there are $K$ parties collaboratively train a machine learning model based on $N$ samples $\{X_i,y_i\}_{i=1}^N$. The data samples $X_i \in \mathbb{R}^{1\times d}$ are feature-partitioned over the participant $K$ parties $\{x_i^k \in \mathbb{R}^{1\times d_k}\}_{k=1}^K$, where $d_k$ represents the feature dimension of party $k$. Without loss of generality, we assume party $K$ holds the labels $Y$. Let $\mathcal{D}_i^{k}\triangleq \{x_i^k\}$ denote the data set of party $k \in \{1,\dots,K-1\}$, $\mathcal{D}_i^{K}\triangleq \{x_i^K, Y_i^K\}$ represent the data set of party $K$ and $\mathcal{D}_i \triangleq \{\mathcal{D}_i^k\}$ be the training samples of all VFL parties. Each party $k$ maintains a neural network model $Net_k$ which is parameterized by weight parameter $w_k$. Then the VFL optimization objective can be formulated as:
\begin{equation}
\mathop{}_{\mathbf{w} }^{\mathbf{min}} \mathcal{L}(\mathbf{w};\mathcal{D}) \triangleq \frac{1}{N} \sum_{i=1}^N \ell(w_1,\dots,w_K;\mathcal{D}_i)
\label{vfl_base_formulation}
\end{equation}
where $\mathbf{w}=\{w_1,\dots,w_K\}$ and $\ell(\cdot)$ is the loss function.

The following presents the formulation  of VFNAS where $K$ parties collaboratively conduct NAS tasks  on the feature-partitioned data samples. Based on the above  formulation, the optimization problem of VFNAS can be formulated as:
\begin{equation}
\mathop{}_{\mathbf{w},\mathcal{A }}^{\mathbf{min}} \mathcal{L}(\mathbf{w},\mathcal{A};\mathcal{D}) \triangleq \frac{1}{N} \sum_{i=1}^{N} \ell(\mathbf{w};\mathcal{A}; \mathcal{D}_i ) 
\label{vfl_NAS_formulation}
\end{equation}
where $\mathcal{A}= \{\alpha_1,\dots,\alpha_K\}$ and $\mathbf{w} = \{w_1,\dots,w_K\}$ represent network architecture parameter composite and the weight parameter composite respectively.

VFNAS is to cooperatively search for optimal network architecture composite $\mathcal{A}$ and optimal weight parameter composite $\mathbf{w}$ across $K$ parties in order to fit the data optimally. During any training or inference process, each party is not allowed to share its raw data.

\section{Methods}\label{sec:methods}
In this section,  we present the basic VFNAS framework  followed by discussions on improving the efficiency and privacy of the basic framework.

\subsection{Vertical Federated Deep Learning}
\begin{figure}[htb]
\center{\includegraphics[width=8cm] {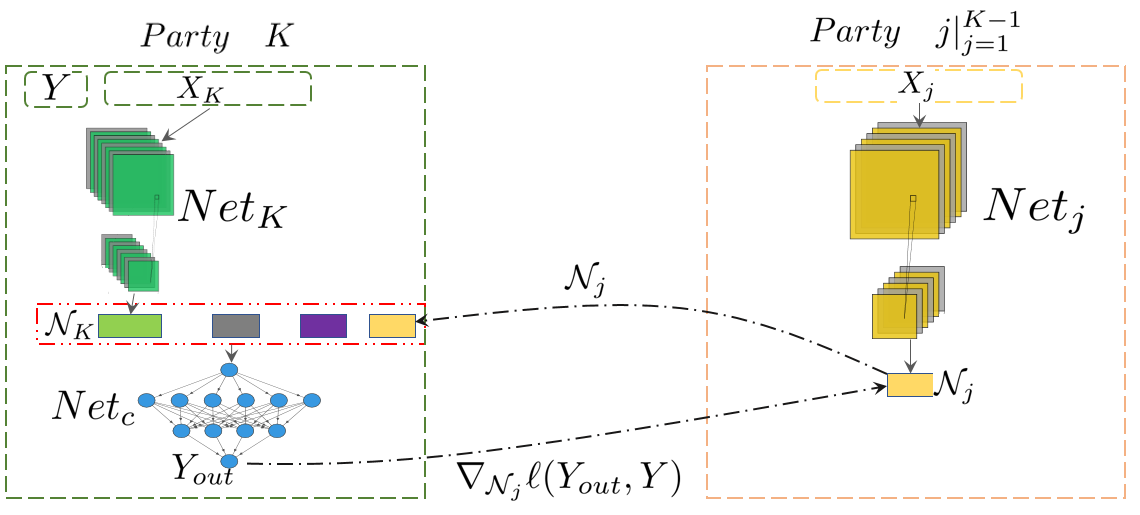}}
\caption{Vertical Federated Deep Learning. Different parties maintain their own network models, which is updated by the exchanges of the intermediate outputs and their corresponding gradients.}
\label{VFNAS_illustration}
\end{figure}

Fig. \ref{VFNAS_illustration} visualizes the basic framework for solving Eqn. \ref{vfl_base_formulation}.
As can be seen in Fig. \ref{VFNAS_illustration}, each party $j$ maintains a network model $Net_j$. Let $\mathcal{N}_j \gets Net_j(w_j;\mathcal{D}_i^j)$ denote the output of $Net_j$. During the training or inference process, each party in $\bigcup_{j=1}^{K-1}j$ has to send the network outputs to party $K$. After receiving $\bigcup_{j=1}^{K-1}\mathcal{N}_j$  from all participant parties, party $K$  merges them accordingly and generates the final classification result, which is conducted on the succeeding network $Net_c$. Let $w_c$, $Y_{out}$ denote the parameter and the final prediction output of $Net_c$, then:
\begin{equation}
Y_{out} \gets Net_c(\mathcal{N}_1,\dots,\mathcal{N}_K;w_c)
\label{yout_formulation}
\end{equation}
Since $w_K$ and $w_c$ are simultaneously updated by party $K$, we reformulated Eqn. \ref{vfl_base_formulation} as 
\begin{equation}
\mathop{}_{\mathbf{w} }^{\mathbf{min}} \mathcal{L}(\mathbf{w},\mathcal{D}) \triangleq \frac{1}{N} \sum_{i=1}^N \ell(w_1,\dots,w_{[K,c]};\mathcal{D}_i)
\label{w_kc_formulation}
\end{equation}
where $w_{[K,c]}$ denotes the weight parameter composite of $Net_K$ and $Net_c$ and $\mathbf{w}=\{w_1,\dots,w_K,w_c\}$. 

The training process works in a forward-and-backward manner as follows:
\begin{enumerate}
\item \textit{Forward step}: Parties $\bigcup_{j=1}^{K-1}j$ generate the intermediate output $\bigcup_{j=1}^{K-1}\mathcal{N}_j$ and send it to party $K$; Based on Eqn. \ref{yout_formulation}, party $K$ computes the output results $Y_{out}$ and loss $\ell(Y_{out},Y_i^K)$ and updates $w_{[K,c]}$ based on back-propagation;
\item \textit{Backward step}: Party $K$ sends the gradients $\bigcup_{j=1}^{K-1}\nabla_{ \mathcal{N}_j}\ell(\cdot)$ to parties $\bigcup_{j=1}^{K-1}j$ respectively, which are used to update $w_j$ based on the chain-rules of back-propagation. 
\end{enumerate}

\subsection{Naive VFNAS Methods}\label{sec:VFNAS-methods}
In this subsection, we  provide the naive VFNAS methods based on two algorithms in literature, i.e., DARTS \cite{liu2018darts:} and MiLeNAS \cite{He2020MiLeNASEN}.

DARTS  formulates NAS tasks as a bi-level optimization problem as:
\begin{equation}
\mathop{\min}_{\alpha}\quad \ell_{val} (w^*(\alpha),\alpha)
\begin{array}{r@{\quad}r@{}l@{\quad}l} 
s.t. \quad w^*(\alpha) = \arg\min_{w} \ell_{trn}(w,\alpha)
\end{array} 
\label{darts_formulation}
\end{equation}
where $\ell_{trn}$ and $\ell_{val}$ is the loss function computed on the train dataset $\mathcal{D}$ and the validation dataset $\mathcal{V}$, which are used to update $w$ and  $\alpha$ respectively. As can be seen from Eqn. \ref{darts_formulation}, the update of $\alpha$ relies on the update of  $w^*({\alpha}) $. Therefore, to implement naive DARTS in a collaborative learning scenario requires exchanging gradient-related information of $\alpha$ and $w$ in a sequential matter among parties per iteration, see Algorithm \ref{VFNAS_algorithm}. To further reduce the communication overhead, we consider a mix-level NAS algorithm, MiLeNAS. 
MiLeNAS \cite{He2020MiLeNASEN} further reformulates Eqn. \ref{darts_formulation} as a mix-level optimization problem by introducing a weight-balancing parameter:
\begin{equation}
\begin{split}
w &= w - \eta_w \nabla_w \ell_{trn}(w,\alpha) \\
\alpha &= \alpha - \eta_ \alpha (\nabla_ \alpha\ell_{trn}(w,\alpha) +\lambda\nabla_ \alpha\ell_{val}(w,\alpha))
\end{split}
\label{milenas_formulation}
\end{equation}
MiLeNAS eliminates the inter-dependent relationships between the optimization processes of  $w$ and  $\alpha$. In a single round, the update of $w$ and $\alpha$ are independent and can be executed in parallel. 

Similarly to existing  gradient-based NAS approaches \cite{liu2018darts:,He2020MiLeNASEN,Cai2019ProxylessNASDN}, we used back-propagation for both the optimizations of   $\mathbf{w}$ and $\mathcal{A}$.  In the following,  we let VFNAS$^1$ represent the DARTS algorithm where the optimization of $\mathcal{A}$ and $\mathbf{w}$ are based on  Eqn. \ref{darts_formulation}, and VFNAS$^2$ denotes the algorithm using MiLeNAS (Eqn. \ref{milenas_formulation}).

\subsection{Privacy Preservation by Differential Privacy}

In this subsection, we introduce differential privacy into VFNAS in order to protect the exchanged messages among parties from revealing essential information. Differential privacy \cite{dwork2006calibrating,dwork2011firm,dwork2014algorithmic} is known for providing theoretical guarantees and quantification of data leakage, where privacy is measured by quantifying the hardness of distinguishing two adjacent databases given certain queries. Its formal definition is given in Definition \ref{definition_1}.

\begin{definition}
A randomized algorithm $\mathcal{M}: \mathcal{X} \rightarrow \mathcal{R}$ with domain $\mathcal{X}$ and range $\mathcal{R}$ achieves ($\varepsilon$, $\delta$)-differential privacy if for all $\mathcal{S} \subseteq \mathcal{R}$ and for any two adjacent databases $\mathcal{D}$ and $\mathcal{D}'$ $\in \mathcal{X}$, it holds that 
\begin{equation}
Pr(\mathcal{M}(\mathcal{D}) \in \mathcal{S}) \leqslant e^{\varepsilon}Pr(\mathcal{M}(\mathcal{D}')\in \mathcal{S}) + \delta.
\end{equation}
\label{definition_1}
\end{definition}

As can be seen from Fig. \ref{VFNAS_illustration}, intermediate outputs from parties $\bigcup_{j=1}^{K-1}j$ are sent to the party $K$ in the forward pass and their corresponding gradients computed at party $K$ are transmitted back to the respective parties for parameter updating in the backward pass. It has been shown that these shared data can lead to the leakage of sensitive information to adversary participants \cite{bhowmick2018protection,zhu2019deep}. To tackle this problem, we adopt the differential privacy with Gaussian mechanism, for both forward and backward passes. In order to prevent the information leakage of $X_j$ in the forward pass, the following presents the basic operations for differential privacy with Gaussian mechanism \cite{dwork2014algorithmic,abadi2016deep}  of $\mathcal{N}_j$:
\begin{equation}
\begin{split}
\bar{\mathcal{N}}_j &\gets \mathcal{N}_j/max(1, \frac{||\mathcal{N}_j||_2}{C_1}) \\
\tilde{\mathcal{N}}_j &\gets \bar{\mathcal{N}}_j + \mathcal{N}(0, \sigma_1^2 C_1^2 \boldsymbol{I})
\end{split}
\label{dp_formulation}
\end{equation}
where $\sigma_1$ represents the noise scale and $C_1$ denotes the norm bound of $\mathcal{N}_j$.    Similarly, in this context, we let  $\sigma_2$ and $C_2$ represent the noise scale and the norm bound for the backward pass of gradients $\nabla\ell$, which is to protect the information of labels.  

We simplify the privacy analysis by considering only communications of $\mathcal{N}_j$ between party $K$ and one of the other parties in $\bigcup_{j=1}^{K-1}j$ in VFNAS$^1$ and point out that the communication of back-propagation messages follows the same privacy analysis. 

For updating $\mathbf{w}$ and $\mathcal{A}$, intermediate network outputs $\mathcal{N}_j$ and gradients $\nabla_{\mathcal{N}_j}\ell$ are clipped and perturbed with a Gaussian random noise. For each party, $\mathcal{D}$ and $\mathcal{V}$ are disjoint, therefore the processes of updating $\mathbf{w}$ and $\mathcal{A}$ can be treated independently for privacy analysis\cite{singh2020differentially}. Moreover, since the update steps of  $\mathbf{w}$ and $\mathcal{A}$ follow the same procedure, in the following analysis we will not distinguish them.

\begin{theorem}
As shown in Eqn. \ref{dp_formulation}, at each step a Gaussian mechanism $\mathcal{M}^{\mathcal{N}_j}$  adds a Gaussian random noise to the output of neural network $\mathcal{N}_j$ from party $j$. This guarantees ($\varepsilon_1, \delta_1$)-differential privacy for each step, if we choose $\sigma_1$ to be
\begin{equation}
\sigma_1 = \frac{\sqrt{2ln(1.25/\delta_1)}}{\varepsilon_1}.
\end{equation}
\end{theorem}

By applying the strong composition theory \cite{dwork2014algorithmic}, we can obtain the overall privacy guarantee of party $\bigcup_{j=1}^{K-1} j$, in training either weight parameters $\mathbf{w}$ or architecture parameters $\mathcal{A}$.

\begin{corollary}
For any $\delta^{\prime}_1 > 0$, the differential privacy scheme in Eqn. \ref{dp_formulation} achieves ($\varepsilon^{\prime}_1$, $T\delta_1+ \delta^{\prime}_1$) differential privacy for mechanism composition $\mathcal{M}^{\mathcal{N}}_{T}$, with
\begin{equation}
\varepsilon^{\prime}_1= \sqrt{2Tln(1/\delta^{\prime}_1)} \varepsilon_1+ T\varepsilon_1(e^{\varepsilon_1} - 1),
\label{iteration_privacy}
\end{equation}
where $T$ is the number of iterations.
\label{corollary}
\end{corollary}

Eqn. \ref{iteration_privacy} means that more training iterations of VFNAS cost more privacy budget.  
Therefore, in order to provide benefits for both communication savings and better privacy guarantees, we further focus on improving the VFNAS communication efficiency.

\subsection{Communication Efficiency and Privacy Improvement by Self-Supervised Learning}\label{subsection:communication_efficiency}

Unlike most of the existing NAS researches that conduct relative approaches on classification tasks,  Self-Supervised NAS (SSNAS) approaches identify good network architectures without data annotations. Liu et al. \cite{Liu2020AreLN}  and Kaplan and Giryes \cite{Kaplan2020SelfsupervisedNA} first studied the performance of self-supervised NAS approaches and showed that even without labels, self-supervised NAS approaches are able to provide reasonably-good network architectures. 
Since only one party has labels in our VFL framework, we consider leveraging SSNAS for local pre-training at all parties before conducting collaborative VFNAS training. We term this strategy \textit{SS-VFNAS}. We point out at least three benefits for SS-VFNAS. First, SS-VFNAS can reduce the communication overhead among parties by requiring less collaboration steps and improve the overall communication efficiency in the VFL systems. Second, privacy is amplified due to less communication. Thirdly but not lastly, the self-supervised pre-training steps of SS-VFNAS allow parties to obtain a neural network that is generalized well across tasks. In the experiments, we will demonstrate these three desiring properties of SS-VFNAS.

In \cite{Kaplan2020SelfsupervisedNA}, the authors introduced SimCLR \cite{Chen2020ASF} to solve SSNAS tasks. However, as presented in the origin paper, the training process of  SimCLR is much  computationally  inefficient, which requires 32 to 128 cloud TPU cores with relatively-large batch size of 4096.   In this work, we employ  MoCoV2 \cite{chen2020improvedbw,he2020Momentumcf}, which has been proven to be much more efficient.  In order to introduce MoCoV2 into the VFNAS framework, we view  $Net_{j}$  as the MoCoV2 encoder and momentum encoder.  MoCoV2 builds representations for high-dimensional inputs such as images by contrastive learning, which are conducted with a dynamic dictionary. MoCoV2 tries to  minimize the InfoNCE loss \cite{Oord2018RepresentationLW},
\begin{equation}
\ell_{\mathbf{w},\mathcal{A}}^{info}(\cdot) = \frac{exp(q\cdot k_{+}/\tau)}{\sum_{i=0}^Kexp(q\cdot k_i/\tau)}
\label{infonce_loss}
\end{equation} 
which minimizes the distance between the positive pairs $q\cdot k_{+}$ and maximizes the distances between $\sum_{i=1}^Kexp(q\cdot k_i/\tau)$. Note that $\tau$ is the temperature, $q\cdot k_0$ is a positive pair which prevents the collapse of the loss.  MoCoV2 conducts substitute process on the InfoNCE optimization by a $(K+1)$-way softmax-based classifier that tries to classify $q$ as $k_{+}$.
The dynamic dictionary is a copy of the encoder that is momentum updated by the following:
\begin{equation}
\theta_k \gets m\theta_k + (1-m)\theta_q, 0<m<1
\end{equation}
where $\theta_q$ and $\theta_k$ denote the parameters of the encoder and the momentum encoder respectively.

In the following, we denote the VFL classification loss Eqn. \ref{vfl_NAS_formulation} as $\ell_{\mathbf{w},\mathcal{A}}^{cls}(\cdot)$. 
As a summary, Algorithm \ref{VFNAS_algorithm} presents the detailed process for SS-VFNAS$^{1}$.  

\begin{algorithm}
\caption{SS-VFNAS with DARTS}
\begin{algorithmic}[1]
\Require train dataset $\mathcal{D}$, validation dataset $\mathcal{V}$,  noise scale $\sigma_1,\sigma_2$,  norm bound $C_1, C_2$ 
\State \textit{Self-Supervised NAS:}
\For{$j$ in $\{1,\dots,K\}$} 
\While{not converged} 
\State    Update $w_j,\alpha_j$ by descending $\ell_{w_j,\alpha_j}^{info}$  (Eqn. \ref{infonce_loss})
\EndWhile
\EndFor
\State
\State \textit{VFNAS:}
\While{not converged} 
\State Update $\mathcal{A}$ by VFNAS-UPDATE($\mathcal{A}$,$\mathcal{V}$)
\State Update $\mathbf{w}$ by VFNAS-UPDATE($\mathbf{w}$,$\mathcal{D}$)
\EndWhile
\State
\Procedure{VFNAS-UPDATE}{$p$,$\mathcal{T}$}
\State \textit{Forward step:}
 \For{Party $j$ in $\{1,\dots,K-1\}$}: 
\State   Compute  $\mathcal{N}_{j} \gets Net_{j}(w_j,\alpha_j;\mathcal{T}^j)$
 \State  Compute  $\tilde{\mathcal{N}}_{j}$ by Eqn. \ref{dp_formulation}  with  norm bound $C_1$, noise scale $\sigma_1$  
\State  Send  $\tilde{\mathcal{N}}_{j}$  to party K
\EndFor
\State
\State \textit{Backward step:}
\State Party K do: 
\State \quad \quad Update $p_K$ by $\ell(p_K,\mathcal{T}^K)$ Eqn. \ref{w_kc_formulation}
\State \For {$j$ in $\{1,\dots,K-1\}$}:   
\State Compute $\nabla_{\tilde{\mathcal{N}}_{j}}\ell$
\State Obtain $\tilde{\nabla}_{\tilde{\mathcal{N}}_{j}}\ell$ by  Eqn. \ref{dp_formulation}  with norm bound $C_2$ and noise scale $\sigma_2$
 \State  Send   $\tilde{\nabla}_{\tilde{\mathcal{N}}_{j}}\ell$ to party $j$
\EndFor
\State 
\State \textit{Finish update:}
 \For{Party $j \in \{1,\dots,K-1\}$} 
\State   Update $p$ by descending $$[\tilde{\nabla}_{\tilde{\mathcal{N}}_{j}}\ell] \nabla_{p_j}{Net}_{j}(w_j;\alpha_j;\mathcal{T}^j)$$
\EndFor
\EndProcedure
\State
\end{algorithmic}
\label{VFNAS_algorithm}
\end{algorithm}

Note that  SS-VFNAS$^2$  can be implemented with slight modifications in Algorithm \ref{VFNAS_algorithm}, which only introduces parallel update processes of $\mathbf{w}$ and $\mathcal{A}$ (referring Eqn. \ref{milenas_formulation}). 

\section{Experiments}\label{sec:experiments}
In this section, we first introduce two datasets for the VFL scenarios by modifying existing datasets. Next, we comprehensively evaluate the performance and efficacy of the SS-VFNAS framework on these datasets. 
Specifically, we compare SS-VFNAS and its variants to several baselines:
\begin{enumerate}
\item Vanilla NAS methods conducted only on party $K$ (which maintains the labels) to demonstrate the superiority of collaborative training over one-party local training, SSNAS. 
\item A VFL framework using established network architectures, including ResNet \cite{he2016deep}, SqueezeNet \cite{iandola2016squeezenet} and ShuffleNet V2 \cite{zhang2018shufflenet}, to show that the proposed SS-VFNAS framework can automate the optimization of the network architectures of different parties simultaneously, while achieving state-of-the-art performance.
\item A VFL framework with naive NAS training without self-supervised pre-training, \textit{VFNAS}. 
\item A VFL framework with end-to-end training combining the self-supervised loss and supervised collaborative training loss, \textit{VFNAS\_E2E}.
\end{enumerate}
Finally, we evaluate the trade-off between privacy and performance by adding various levels of privacy budget in SS-VFNAS.

\subsection{Experimental Settings}

\subsubsection{Dataset}\label{dataset}
Despite the fast growth of research effort on FL, there is still lack of real-world vision dataset and benchmarks for VFL scenarios. Existing works either consider cross-modal problems (i.e., image-text) which limits to two parties or artificially partition one image into multiple segments to simulate multi-party scenarios. In this subsection, we introduce two benchmark datasets that are used for our evaluation, ModelNet40 \cite{Wu20153D} and CheXpert-14 \cite{Irvin2019CheXpertAL}, with tailored modifications to mimic the real-world data distribution. 

\begin{figure*}[htb]
\center{\includegraphics[width=12cm] {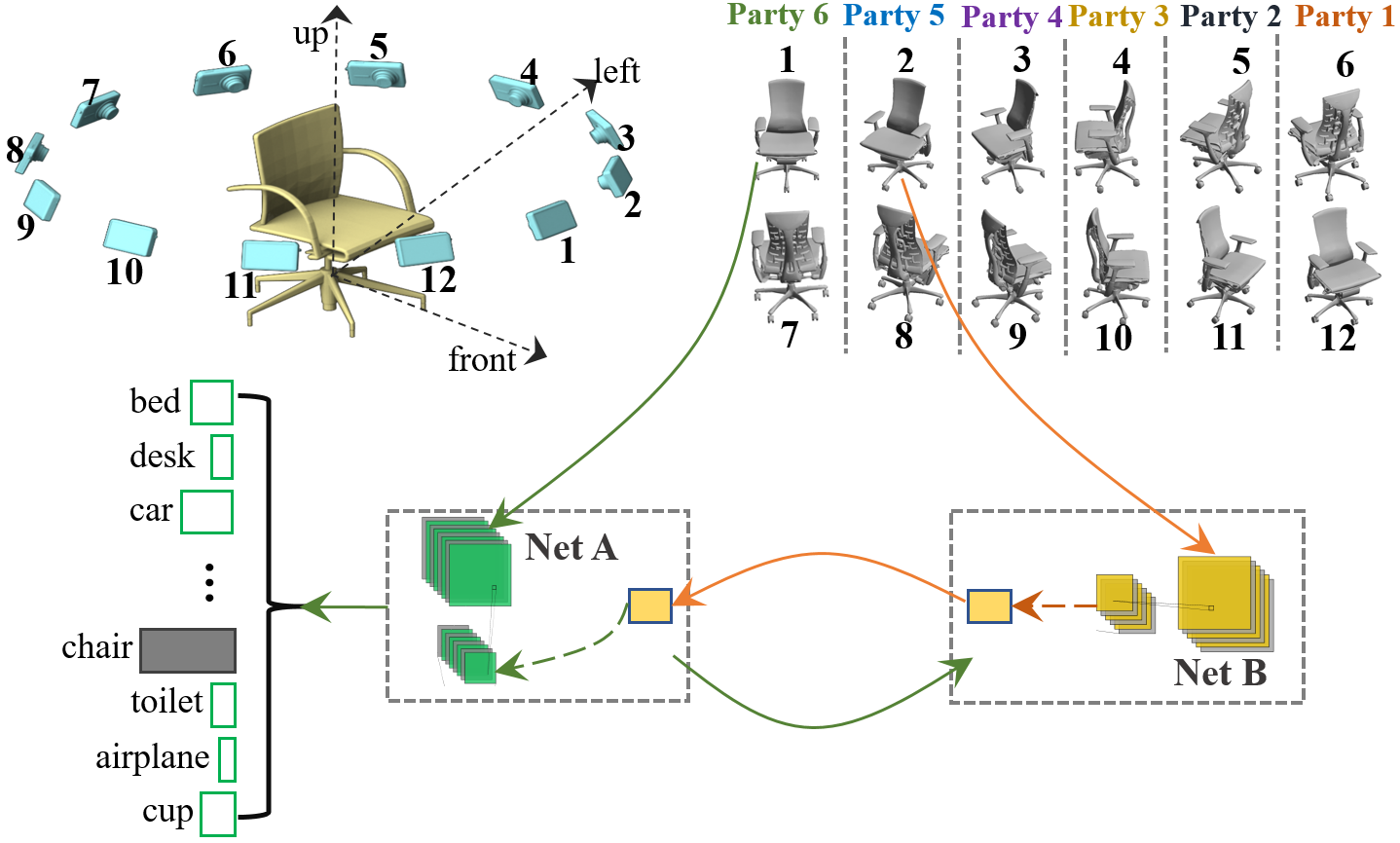}}
\caption{ The process for reforging ModelNet40 for   VFL classification benchmark. Firstly, we generate multi-view images based on the approaches in \cite{Su2015Multi}. Then we distribute the images evenly to different parties. Each VFL sample is generated by taking one single-view image from each party in sequence.}
\label{dataset_reforge}
\end{figure*}

\noindent\textbf{FedModelNet40}\label{dataset_reforging}
ModelNet is a widely-used 3D shape classification and shape retrieval benchmark, which currently contains 127,915 3D CAD models from 662 categories. We use a 40-class well-annotated subset containing 12,311 shapes from 40 common categories, ModelNet40\footnote{http://modelnet.cs.princeton.edu/}. 
We created 12 2D multi-view images per 3D mesh model by placing 12 virtual cameras evenly distributed around the centroid and partitioned the images into multiple (2 to 6) parties by their angles. The classification labels are allocated to party K. Fig. \ref{dataset_reforge} is an illustration of the dataset generation process. The training dataset are split into train, validation and test dataset containing 3183, 3183 and 1600 samples. We term this dataset \textit{FedModelNet40}. 

\noindent\textbf{FedCheXpert}
CheXpert-14 \cite{Irvin2019CheXpertAL} is a large dataset for  chest radiograph interpretation, which includes 224,316 chest radiographs of 65,240 patients.  
We partition the radiograph into two parties: the front view and the lateral view and consider 5 classification challenges, including Cardiomegaly, Edema, Consolidation, Atelectasis and Pleural Effusion\footnote{https://stanfordmlgroup.github.io/competitions/chexpert/}. For each classification task, we place corresponding classification head,  which is independently connected to the CNN-based backbones. Then the total loss function is calculated as the summation of the losses computed on each task. The dataset contains 11252 training samples, 11252 validation samples and 6126 test samples. We name this dataset \textit{FedCheXpert}.


\subsubsection{ Architecture Search Space}
We adopted the architecture search space used in DARTS and MiLeNAS, which has been proven to be efficient in many NAS tasks.  DARTS constructs two different convolution cells to build the entire network structure. Each cell contains different nodes connected by different operations (e.g., convolution, max pooling, skip connection, zero). Let $\mathcal{O}$ represent the operation set between node $(i,j)$, to make the search space differentiable, DARTS makes softmax relaxation over all possible operations between $(i,j)$:
\begin{equation}
\bar{o}^{(i,j)} = \sum_{o \in \mathcal{O} } \frac{exp(\alpha_{o}^{(i,j)})}{\sum_{o^{\prime}\in \mathcal{O}}exp(\alpha_{o^{\prime}}^{(i,j)})}o(x)
\end{equation}
where the architecture weights is parameterized by searching over the differentiable vector space $\alpha^{(i,j)}$.  

\subsubsection{Parameter Settings}



In the search process, the network outputs $\bigcup_{j=1}^K\mathcal{N}_j$ are the visual representations in MoCoV2, protected by various levels of noise.  Each $\bigcup_{j=1}^K\mathcal{N}_j$ is set 64-dimensional. 
Other parameter settings are introduced from the original paper of MiLeNAS \cite{He2020MiLeNASEN} and DARTS\cite{he2020Momentumcf}.

In the architecture evaluation process,  we adjust the number of stacked layers of the searched architectures in order to generate and evaluate various architecture complexity and performance, including SS-VFNAS-S with 4 layers,   SS-VFNAS-M with 8 layers and SS-VFNAS-L with 14 layers.  The classification net ${Net}_c$ contains two fully-connected layers with 512 and 128 neurons, using $tanh$ activation function. 

For the manually-designed network, we replace ${Net}_j$ with the corresponding networks. We keep the network settings in ${Net}_c$ in order to make fair comparisons.  In order adjust to the NVIDIA GTX 2080Ti GPU used, we fix the batch size to be 32. 



\subsection{Experiment Results}

\subsubsection{Effectiveness of SS-VFNAS}

Table \ref{nas_comp_result}  presents the Top-1 and Top-5 accuracy results of SS-VFNAS variants and some existing popular networks, including ResNet \cite{he2016deep}, SqueezeNet \cite{iandola2016squeezenet} and ShuffleNet V2 \cite{zhang2018shufflenet} obtained on a two-party VFL experiments on FedModelNet40. Table \ref{chexpert_compare} is the accuracy and AUC results obtained on different tasks in  FedCheXpert.

\begin{table*}[htbp] 
\footnotesize
\centering
\caption{Test Accuracy on FedModelNet40 of SS-VFNAS variants and some popular backbone architectures, including ResNet, SqueezeNet and ShuffleNet V2. \#P and \#F represent the parameter size (MB) and the FLOPS(M). The results in bold denote the best ones obtained in the corresponding Top-1 or Top-5 accuracy. }
\begin{tabular}{cccccc|cccccc}
\toprule[2.0pt]
Model & Type & \#P & \#F & Top-1(\%) & Top-5(\%) & Model & Type & \#P & \#F & Top-1(\%) & Top-5(\%)\\ 
\cmidrule[1.2pt](r){1-12}
ResNet18 & manual & 11.21& 1818 & 81.87 & 95.25 & SSNAS-S & Auto & 1.18   & 183  & 81.88  & 95.24 \\ 
ResNet34 & manual & 21.31& 3670 & 81.31 & 95.25 & SSNAS-M & Auto &  2.35 & 340  &81.81   & 95.13 \\ 
ResNet50 & manual & 23.64& 4109 & 81.18 & 95.62 & SSNAS-L & Auto &  3.88 & 513  &81.18   &  94.97\\ 
ShuffleNet V2 & manual & 1.32 & 147 & 79.00 & 94.56\\ 
SqueezeNet & manual & 0.77 & 742 & 78.94 & 95.63\\ 

\cmidrule[1.2pt](r){1-12}
VFNAS$^{1}$-S & Auto &1.06 & 161 & 81.43 & 95.68  & VFNAS$^{2}$-S & Auto & 1.07 &  172 & 82.38  & 95.40 \\ 
VFNAS$^{1}$-M & Auto  & 2.25 & 309 & 82.32 & 95.94 & VFNAS$^{2}$-M & Auto &  2.32 & 321  &  82.31 &  95.45\\ 
VFNAS$^{1}$-L & Auto & 4.21 & 517 & 82.00 & 96.50$^{\dagger}$ & VFNAS$^{2}$-L & Auto &   3.88  & 519  & 81.43 & 95.07 \\ 

\cmidrule[1.2pt](r){1-12}
SS-VFNAS$^{1}$-S & Auto & 1.12  & 172   & 82.79  & 95.64  & SS-VFNAS$^{2}$-S & Auto & 1.33 &  193  & 81.75  &  95.40 \\ 
SS-VFNAS$^{1}$-M & Auto &  2.29 & 332  & \textbf{83.88}$^{\star}$ &  \textbf{97.12}$^{\dagger}$ & SS-VFNAS$^{2}$-M & Auto & 2.91  & 330   &  82.81$^{\star}$  & 96.37 \\ 
SS-VFNAS$^{1}$-L & Auto &  4.72 &  539 &  83.01$^{\star}$ & 94.12  & SS-VFNAS$^{2}$-L & Auto &  4.54 &  531  &  82.19& 96.53$^{\dagger}$ \\

\bottomrule[1.8pt] 
\end{tabular}
\label{nas_comp_result}
\end{table*}

As can be seen in Table \ref{nas_comp_result} and \ref{chexpert_compare}, SS-VFNAS and its variants are efficient in finding promising architectures. Compared with  ResNet, SqueezeNet and ShuffleNet V2, SS-VFNAS can reach comparable or superior performance with much smaller network architectures. 

\begin{table*}[htbp] 
\footnotesize
\centering
\caption{Test Accuracy  on FedCheXpert, including SS-VFNAS variants and some popular backbone architectures. \#P  represents the parameter size (MB) of the corresponding backbones. The results in bold denote the best one obtained in the corresponding test  accuracy. Mean represents the average test accuracy or AUC  results. }
\begin{tabular}{ccc|cc|cc|cc|cc|cc|cc}
\toprule[2.0pt]
\multirow{2}{*}{\textbf{Model}}  & \multirow{2}{*}{\textbf{Type}} & \multirow{2}{*}{\#P} & 
        \multicolumn{2}{c}{\textbf{Cardiomegaly}} & \multicolumn{2}{c}{\textbf{Edema}} & \multicolumn{2}{c}{\textbf{Consolidation}} & \multicolumn{2}{c}{\textbf{Atelectasis}} & \multicolumn{2}{c}{\textbf{Pleural Effusion}} & \multicolumn{2}{c}{\textbf{Mean}}  \\
        \cmidrule{4-15}
        &&&  \textbf{Acc} & \textbf{AUC}  &  \textbf{Acc} & \textbf{AUC}  &  \textbf{Acc} & \textbf{AUC} & \textbf{Acc} & \textbf{AUC} &  \textbf{Acc} & \textbf{AUC}  & \textbf{Acc} & \textbf{AUC}  \\ \midrule
        ResNet18 & Mannual & 11.2  &0.904 &0.838 &	0.896 	&0.848 	&0.942 	&0.698 	&0.771 	&0.723 	&0.830 	&0.878  &0.869 	&0.797  \\
        ResNet34 & Mannual & 21.3&0.901  	&0.836 	& \textbf{0.899} 	&0.843 	&0.942 	&0.708 	&0.772 	&0.720 	&0.829 	&0.873  &0.869 	&0.796  \\
        ResNet50 & Mannual & 23.6&0.902 	&0.817 	&0.895 	&0.830 	&\textbf{0.943} 	&0.686 	&0.769 	&0.708 	&0.820 	&0.858 & 0.866 	&0.780  \\
        ShuffleNet V2 & Mannual& 1.3 & 	0.902 	&0.823 	&0.894 	&0.835 	&\textbf{0.943} 	&0.683 	&0.772 	&0.703 	&0.814 	&0.857 &0.865 	&0.780 \\
        SqueezeNet & Mannual & 0.8 &0.903 &	0.829 	&0.897 	&0.840 	&\textbf{0.943} 	&0.696 &	0.769 	&0.710 	&0.813 	&0.849 &0.865 	&0.785 \\
\cmidrule{1-15}
        SSNAS-M & Auto& 2.4 & 0.900 	&0.846 	&0.889 	&0.846 	&0.941 	& \textbf{0.724} 	&0.771 	&0.726 	&0.830 	&0.883  &0.866 	&\textbf{0.805}  \\
\cmidrule{1-15}
        SSNAS\_E2E-M & Auto& 2.3 & 0.902 	&0.834 	&0.893 	&0.838 	&\textbf{0.943} 	& 0.698 	&0.770 	&0.713 	&0.823 	&0.880  &0.866 	&0.793  \\
\cmidrule{1-15}
        VFNAS$^{1}$-M & Auto& 2.4 &0.903 	&0.844 	&0.893 	&0.842 	&\textbf{0.943} 	&0.705 	&0.770 	&0.720 	&0.838 	&0.883  &0.869 	&0.799  \\
        VFNAS$^{2}$-M & Auto & 2.4 &0.903 	&0.842 	&0.893 	&0.844 	&0.942 	&0.709 	&0.772 &0.718 	&\textbf{0.840} &	0.889 & 0.870 	&0.800  \\
\cmidrule{1-15}
        SS-VFNAS$^{1}$-M & Auto &2.3 & \textbf{0.907} 	&\textbf{0.848} 	&0.898 	&0.846 &	\textbf{0.943} 	&0.716 	&\textbf{0.779} 	&\textbf{0.727} 	&0.839 	&\textbf{0.890}  &\textbf{0.873} 	&\textbf{0.805}  \\
        SS-VFNAS$^{2}$-M & Auto& 2.4  & 0.903 &	0.840 &	0.897 	& \textbf{0.849} 	&\textbf{0.943} 	&0.702 	&0.776 &	\textbf{0.727} 	&0.836 	&0.889 & 0.871 	&0.802  \\
\bottomrule[1.8pt] 
\end{tabular}
\label{chexpert_compare}
\end{table*}

\begin{figure*}[!t]
\centering
\subfloat[Model performance on FedModelNet40.]{\includegraphics[width=2.75in]{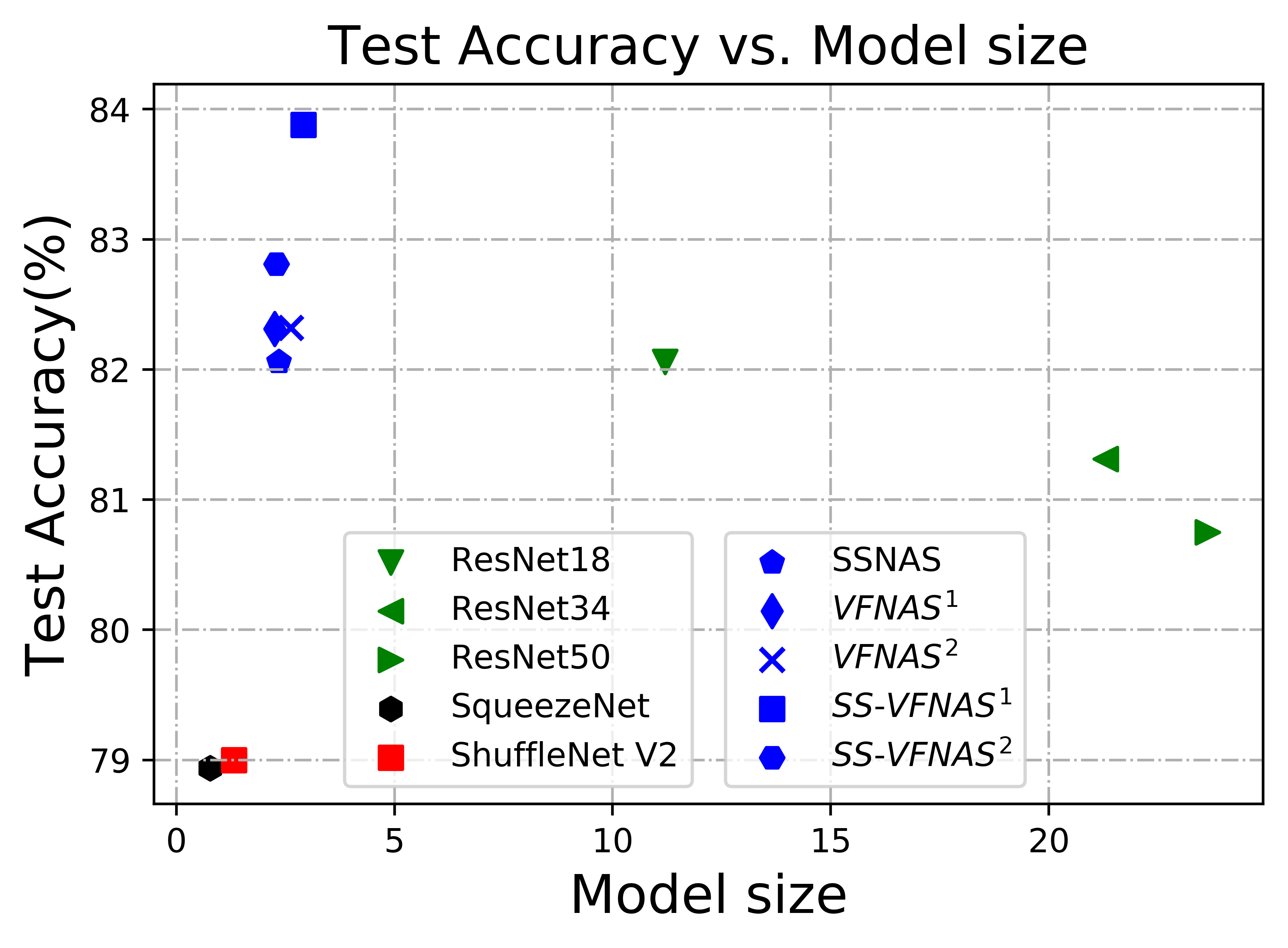}}
\subfloat[Model performance on  FedCheXpert.]{\includegraphics[width=2.75in]{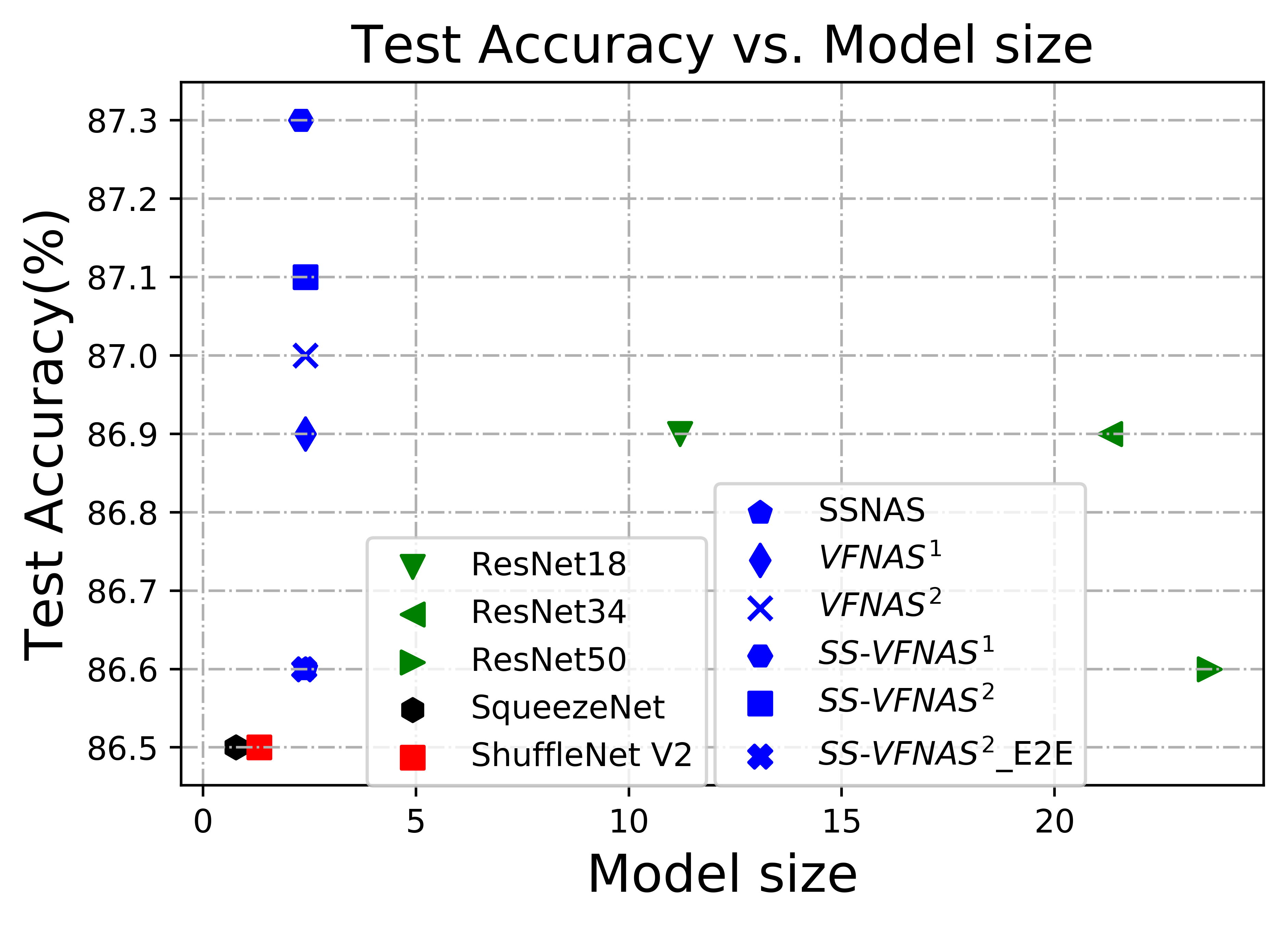}}
\caption{Test Accuracy VS Model Size.}
\label{model_size_cmpare}
\end{figure*}

\begin{figure}[!t]
\centering
\subfloat[Normal Cell of Party 5]{\includegraphics[width=2.7in]{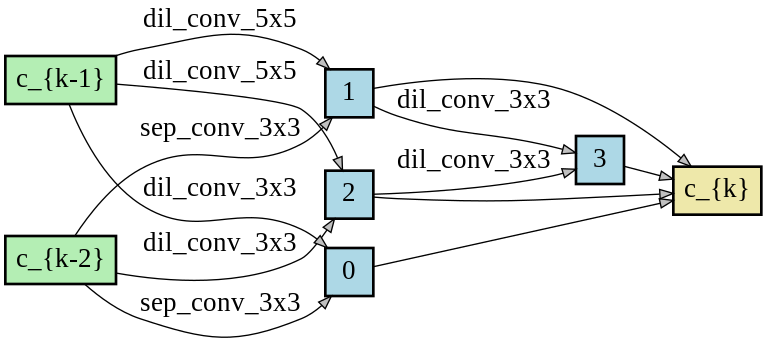}}
\label{Areduce}
\subfloat[Reduction Cell of Party 5]{\includegraphics[width=2.7in]{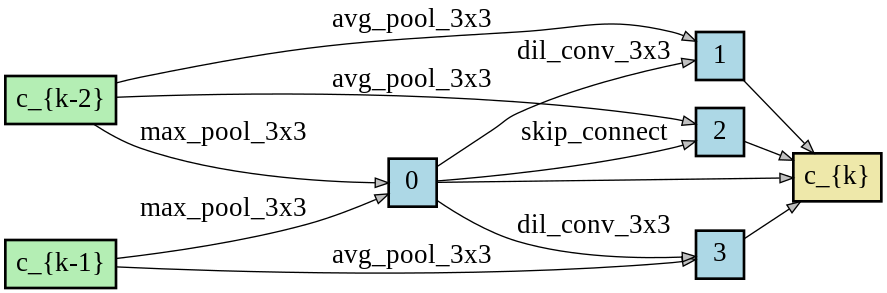}}
\label{Breduce}
\subfloat[Normal Cell of Party 6]{\includegraphics[width=2.7in]{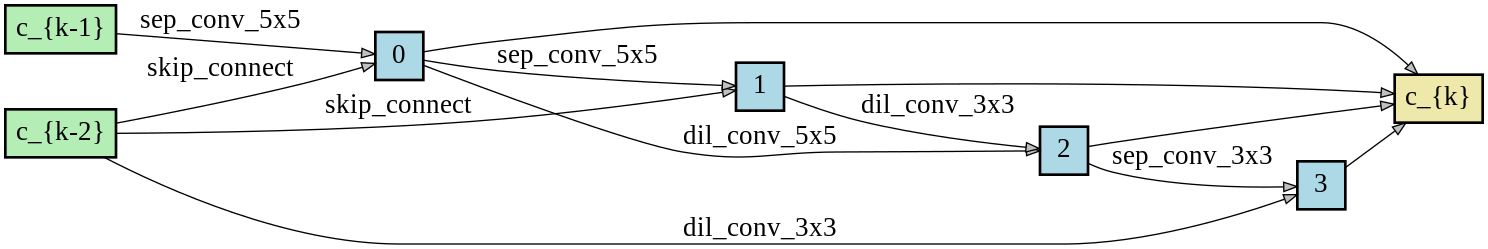}}
\label{Anormal}
\subfloat[Reduction Cell of Party 6]{\includegraphics[width=2.7in]{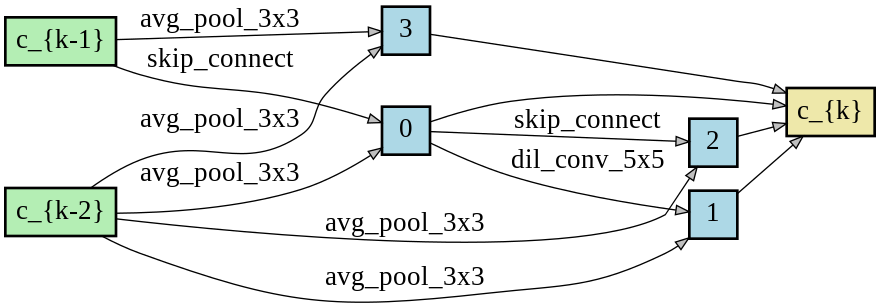}}
\label{Bnormal}
\caption{Optimal network architecture searched by SS-VFNAS$^{1}$-M with parties 6 and 5 in FedModelNet40.}
\label{Optimal_architecture}
\end{figure}

Notice that the architectures stacked the most layers including SSNAS-L, VFNAS-L, SS-VFNAS-L do not always achieve the best results, indicating that bigger network architectures are not necessarily the best choices. In order to better illustrate the model efficiency, we present Fig. \ref{model_size_cmpare}, which compare various models from the perspectives of both accuracy and model size. It can be observed that with the ability of simultaneously searching for the optimal architectures for different parties, SS-VFNAS obtains the best model accuracy and efficiency among all algorithms, followed by VFNAS-type algorithms which achieve slightly lower accuracy but also compact architectures. Fig. \ref{Optimal_architecture} visualizes the differences in the optimal architectures obtained by SS-VFNAS$^{1}$-M between the two collaborating parties (party 5 and 6) on FedModelNet40. We observe from Fig. \ref{Optimal_architecture} that the optimal architectures for different parties may not necessarily be the same, even though the image data are from different angles of the same object, which demonstrates that manually designing identical networks for different parties is not the optimal choice.

\subsubsection{The Impact of Number of Parties}
In this section, we show the essentiality of SS-VFNAS by answering the following question: from the perspective of party K, is it more beneficial to cooperate with some of the parties in $\bigcup_{j=1}^{K-1}j$ within SS-VFNAS than simply carrying out classification task using its own data? 

\begin{figure*}[!t]
\centering
\subfloat[Model performance on FedModelNet40.]{\includegraphics[width=2.75in]{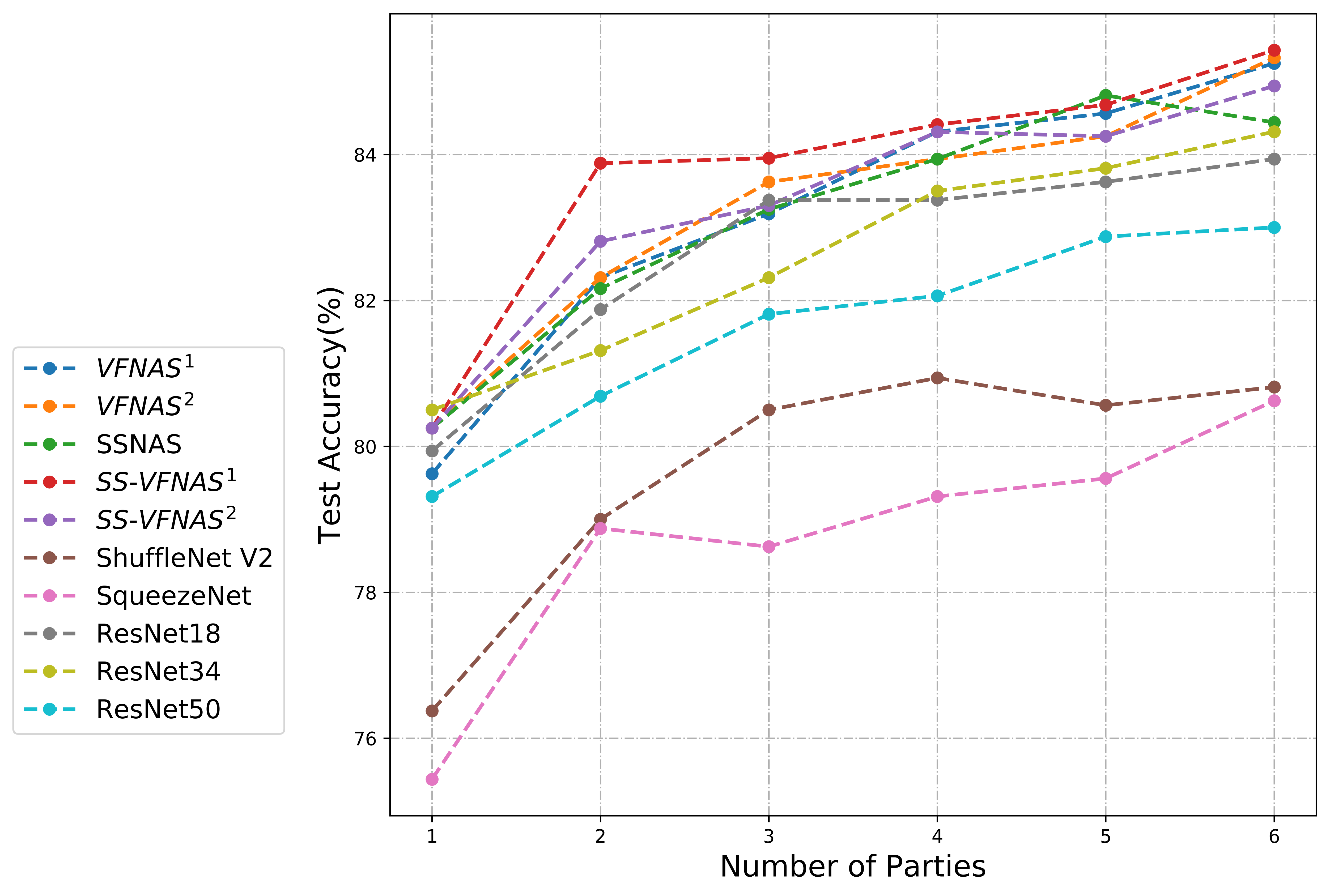}}
\subfloat[Model performance on  FedCheXpert.]{\includegraphics[width=2.75in]{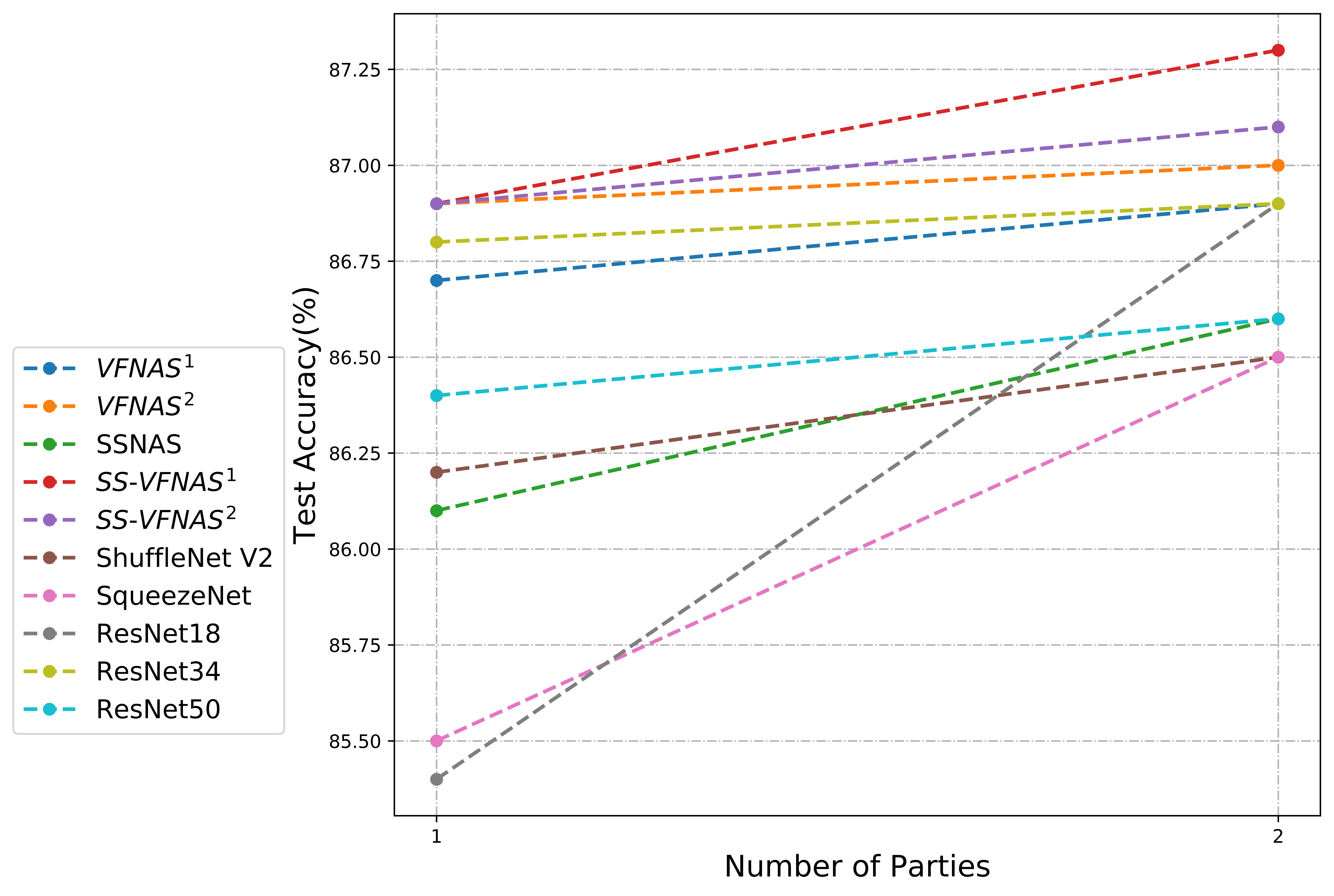}}
\caption{Test Accuracy for different number of participant parties on FedModelNet40 and FedCheXpert.} 
\label{vflnasessenntiality}
\end{figure*}

We evaluate the impact of number of parties using various network backbones, i.e.,  ResNet \cite{he2016deep}, SqueezeNet \cite{iandola2016squeezenet} and ShuffleNet V2 \cite{zhang2018shufflenet}, as well as automated approaches including VFNAS$^{1}$, VFNAS$^{2}$ and SS-VFNAS$^{1}$ and SS-VFNAS$^{2}$. 
After the architecture search process on the training and validation dataset is completed, the corresponding architecture is evaluated on the test dataset. Fig. \ref{vflnasessenntiality} presents the test accuracy obtained by different algorithm settings.

As can be seen from Fig. \ref{vflnasessenntiality}, the test accuracy is gradually improved by adding more participating  parties, which is also consistent with the conclusions obtained in \cite{Su2015Multi}, demonstrating 
that SS-VFNAS is capable of extracting complementary information of the multi-view images distributed among parties $\{1,\dots,K\}$. In addition, we observe that the performance of SS-VFNAS and its variants is consistently higher than other methods, followed by VFNAS-algorithms, demonstrating the effectiveness of our proposed SS-VFNAS. Since more parties involved means higher computation and communication complexity, for the rest of our experiments, we conduct the two-party experiments on FedModelNet40 dataset with the participant of parties 5 and 6.

\subsubsection{Communication Efficiency}\label{subsection:commu_efficiency}

Although gradient-based NAS approaches have been proven to be relatively computational efficient, they are still very expensive to achieve state-of-the-art performance. 
If multiple parties perform collaborative NAS training, it will require prohibitive communication overhead. One of the most critical goals of VFNAS is to obtain reasonable performance with as few communication rounds as possible. 

In this section, we aim to evaluate the communication efficiency of different SS-VFNAS algorithms on the FedModelNet40 dataset, including  VFNAS$^{1}$-M, VFNAS$^{2}$-M, SS-VFNAS$^{1}$-M,  SS-VFNAS$^{2}$-M and VFNAS\_E2E-M, where VFNAS\_E2E-M represents  an end-to-end training approach which parties collaboratively train a model with an objective that combines the unsupervised and supervised loss as:
\begin{equation}
\mathop{}_{\mathbf{w},\mathcal{A}}^{\mathbf{min}} \ell_{\mathbf{w},\mathcal{A}}^{cls}+ \gamma \ell_{\mathbf{w},\mathcal{A}}^{info}, \gamma>0
\label{end2end_loss}
\end{equation}
In the following evaluation, the hyper-parameter $\gamma$ is set to be $0.1$.

\begin{figure}[!t]
\centering
\subfloat[Training Loss.]{\includegraphics[width=1.75in]{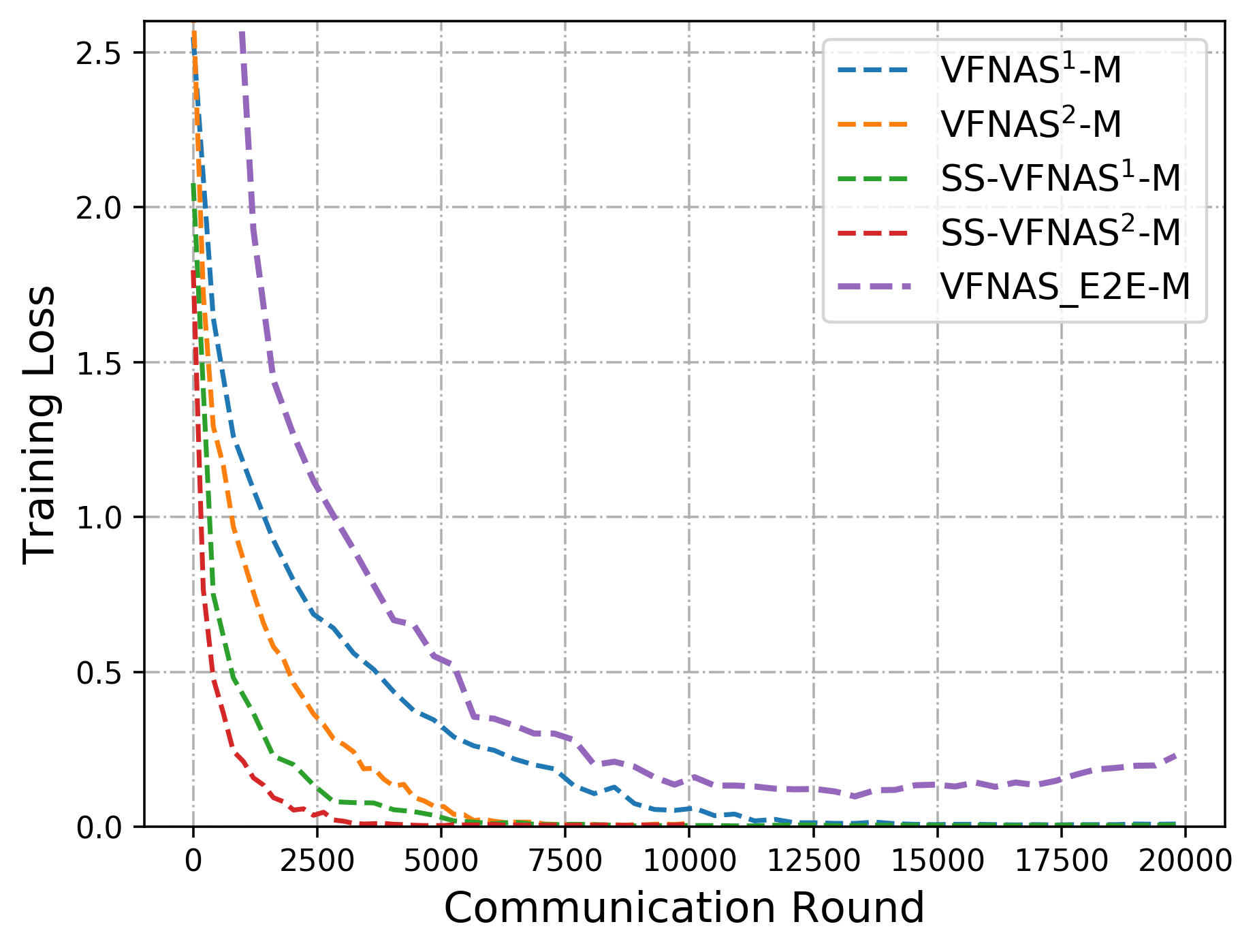}}
\subfloat[Validation Accuracy.]{\includegraphics[width=1.75in]{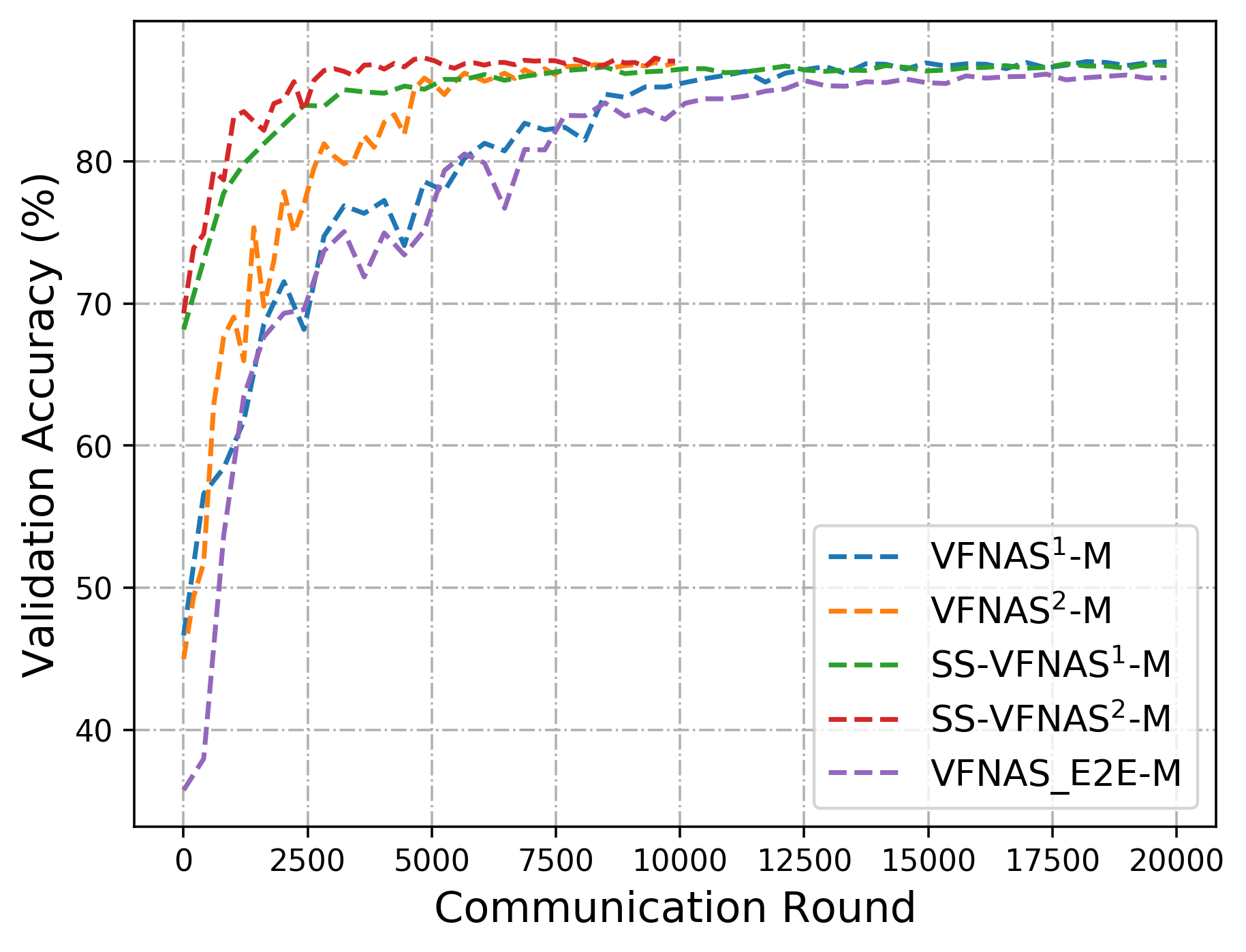}}
\caption{Communication efficiency of various algorithms for a two-party VFL setting on FedModelNet40}
\label{communication_compare}
\end{figure}

Fig. \ref{communication_compare} visualizes the training loss and the validation accuracy of SS-VFNAS variants with respect to the communication rounds in a two-party VFL. Table \ref{communication_table} presents the total communication rounds and time consumed, as well as the model size (the number of parameters) to converge. In order to avoid stochastic deviation, we view each algorithm converged only if its validation accuracy stops improving after more than 5 iterations.

\begin{table}[htbp] 
\footnotesize
\centering
\caption{The communication rounds that different algorithms require to converge.}
\begin{tabular}{cccc}
\toprule[2.0pt]
Model & Parameter size(MB) & Communication\_round  \\ 
\cmidrule[1.2pt](r){1-3}
VFNAS$^{1}$-M &2.25& 11484 \\ 
VFNAS$^{2}$-M &2.32& 5742 \\ 
SS-VFNAS$^{1}$-M &2.29& 8712 \\ 
SS-VFNAS$^{2}$-M &2.91& 4950 \\ 
VFNAS\_E2E-M &2.34& 17424 \\ 

\bottomrule[1.8pt] 
\end{tabular}
\label{communication_table}
\end{table}

From Fig. \ref{communication_compare} and Table \ref{communication_table}, the following observations are made:
\begin{enumerate}
\item  VFNAS$^{2}$-M and SS-VFNAS$^{2}$-M  need about half of the communication rounds that VFNAS$^{1}$-M and SS-VFNAS$^{1}$-M need to reach the same accuracy level, indicating that the parallel execution of the update  of $\mathbf{w}$ and $\mathcal{A}$ can speed up the training process, as explained in subsection \ref{sec:VFNAS-methods} in detail.
\item SS-VFNAS$^{1}$-M and SS-VFNAS$^{2}$-M require much less communication rounds than VFNAS$^{1}$-M and VFNAS$^{2}$-M, respectively, demonstrating that executing VFNAS as a downstream task of SSNAS can greatly improve its communication efficiency; On the other hand, VFNAS\_E2E-M requires the most communication round to converge and appears to reach a lower accuracy, possibly because employing an end-to-end loss like Eqn. \ref{end2end_loss} adversely affect the training towards the supervised learning objective.
\end{enumerate}

\subsubsection{Learning with Limited Overlapping Samples}
\begin{figure}[htb]
\center{\includegraphics[width=8.6cm] {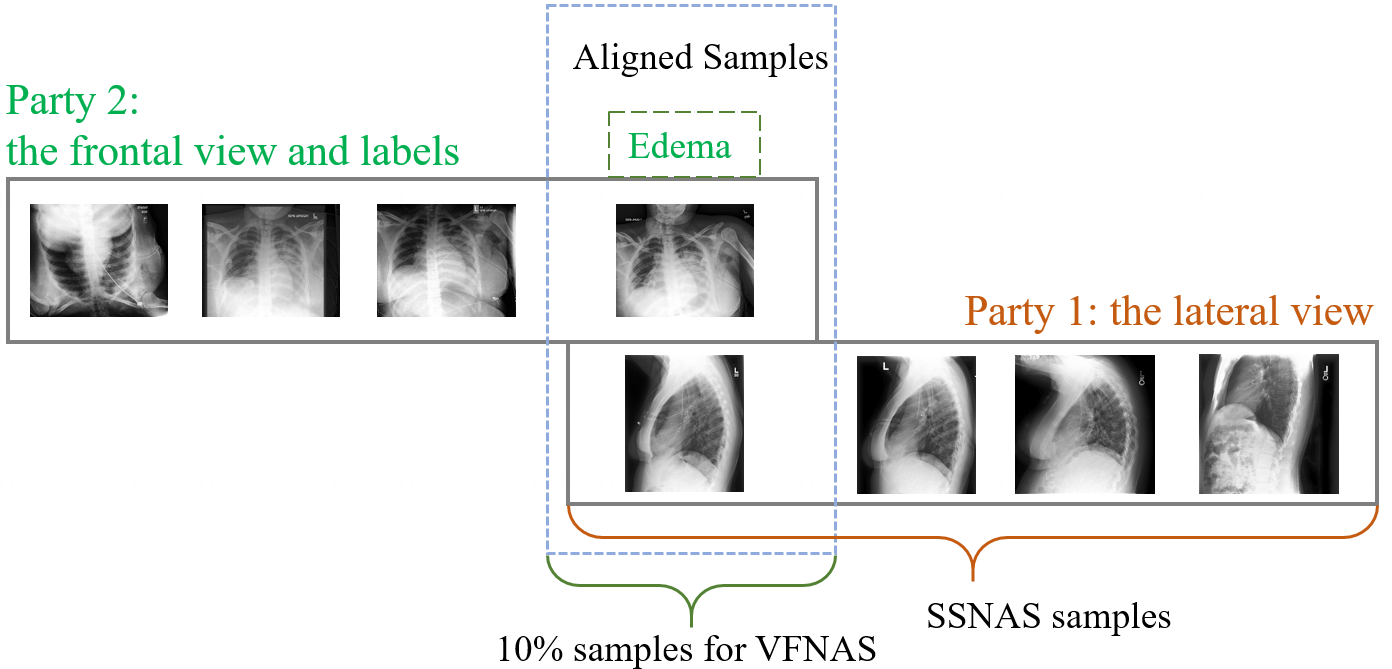}}
\caption{A two-party VFL setting with only 10\% aligned FedCheXpert samples.}
\label{transfer_align_samples}
\end{figure}

In VFL scenarios, a prerequisite is to have sufficient overlapping samples and labels with distributed features. However labels are often expensive to obtain, and as the number of parties grow in VFL, the number of overlapping samples may decrease dramatically. Unsupervised or self-supervised training can help to eliminate the dependency on labels or large number of overlapping samples. One of the advantages of performing self-supervised learning at each local party before conducting federated learning is the transferability of the pre-trained architectures. For a K-party federation, we first obtain K self-learned representations from each party's data, respectively. Then we use these representations to perform various downstream tasks. This can greatly improve the scalability of VFL tasks. As we demonstrated below, even using representations self-learned from different domains, parties can still perform downstream VFL tasks with higher accuracy than using only limited data from only the task domains. Here, we consider a scenario where only limited overlapping samples (90\% less than previous experiments) and labels are available in a two-party VFL whereas each party holds a large number of non-overlapping non-labeled data, see Fig \ref{transfer_align_samples}. Note such a scenario has also been proposed in previous work for the study of FTL \cite{Liu2020ASF}. Specifically, we studied the following settings : 
\begin{itemize}
\item SS$^{c+c}$-VFNAS$^1$. Each party performs self-supervised learning with its respective samples from FedCheXpert. Then with the learned representations, they perform VFNAS on the 10\% overlapping samples (Fig. \ref{transfer_align_samples});
\item SS$^{c+m}$-VFNAS$^1$. Similar to the previous setting, except that only one party holds samples from FedCheXpert and the other holds samples from FedModelNet40, with which they pre-train self-supervised representations and architectures. 
\end{itemize}

We compared the performance of the above settings to that of the settings with only the limited overlapping samples but no other alternative data, with and without self-supervised training. The results are shown in Table \ref{few_shot_chexpert_compare}. From Table \ref{few_shot_chexpert_compare}, we observe that the performance of SS-VFNAS$^1$ and VFNAS$^1$ are not as good as its counterparties in Table \ref{chexpert_compare} due to the reduction of training samples; However, both SS$^{c+c}$-VFNAS$^1$ and SS$^{c+m}$-VFNAS$^1$ 
outperform SS-VFNAS$^1$, by a greater margin than that of the case with 100\% overlapping samples as in Table \ref{chexpert_compare}, showing that the benefit of SS-VFNAS is more pronounced as the number of overlapping samples decrease. In addition, the performance of SS$^{c+m}$-VFNAS$^1$ is comparable to SS$^{c+c}$-VFNAS$^1$, which demonstrates that the architectures and representations obtained by self-supervised learning can provide sufficiently-good initial architectures for downstream tasks, even when the pre-trained data set at some parties is from a different domain.

\begin{table*}[htbp] 
\footnotesize
\centering
\caption{Evaluations of  SS-VFNAS$^1$, SS$^{c+c}$-VFNAS$^1$ and SS$^{c+m}$-VFNAS$^1$ on FedCheXpert with 10\% overlapping samples.  The results in bold denote the best one obtained in the corresponding test  accuracy.} 
\begin{tabular}{cc|cc|cc|cc|cc|cc|cc}
\toprule[2.0pt]
\multirow{2}{*}{\textbf{Model}}   & \multirow{2}{*}{\#P} & 
        \multicolumn{2}{c}{\textbf{Cardiomegaly}} & \multicolumn{2}{c}{\textbf{Edema}} & \multicolumn{2}{c}{\textbf{Consolidation}} & \multicolumn{2}{c}{\textbf{Atelectasis}} & \multicolumn{2}{c}{\textbf{Pleural Effusion}} & \multicolumn{2}{c}{\textbf{Mean}}  \\
        \cmidrule{3-14}
        &&  \textbf{Acc} & \textbf{AUC}  &  \textbf{Acc} & \textbf{AUC}  &  \textbf{Acc} & \textbf{AUC} & \textbf{Acc} & \textbf{AUC} &  \textbf{Acc} & \textbf{AUC}  & \textbf{Acc} & \textbf{AUC}  \\ \midrule
        VFNAS$^1$ & 2.4 & 0.886 	&0.702 &	0.887 & 	0.760  & 	0.942 & 	0.622 &	0.765 	& 0.636 &	0.761 	&0.772&  0.848 	&0.698 \\
\cmidrule{1-14}
        SS-VFNAS$^1$ & 2.4 & 0.894 &	0.774& 	0.887 &	0.791& 	0.942 &	0.624 	&0.770 &	0.676 	&0.782 &	0.802 & 0.855 &	0.733  \\
\cmidrule{1-14}
        SS$^{c+c}$-VFNAS$^1$ & 2.4 & \textbf{0.897} 	&\textbf{0.783} &	\textbf{0.888} 	&0.793 &	\textbf{0.943} 	&0.635 &	0.769 &	0.682& 	\textbf{0.783} &	\textbf{0.807} & \textbf{0.856 }	&0.740  \\
\cmidrule{1-14}
       SS$^{c+m}$-VFNAS$^1$ & 2.3 &0.896  &	0.776 	&\textbf{0.888} 	&\textbf{0.803} &	\textbf{0.943} 	&\textbf{0.639} &	\textbf{0.771} 	&\textbf{0.691} &	0.780 	&0.802&  \textbf{0.856}	&\textbf{0.742}   \\
\bottomrule[1.8pt] 
\end{tabular}
\label{few_shot_chexpert_compare}
\end{table*}



\subsubsection{Performance and Privacy Level Trade-off}
\begin{figure}[htb]
\center{\includegraphics[width=7.6cm] {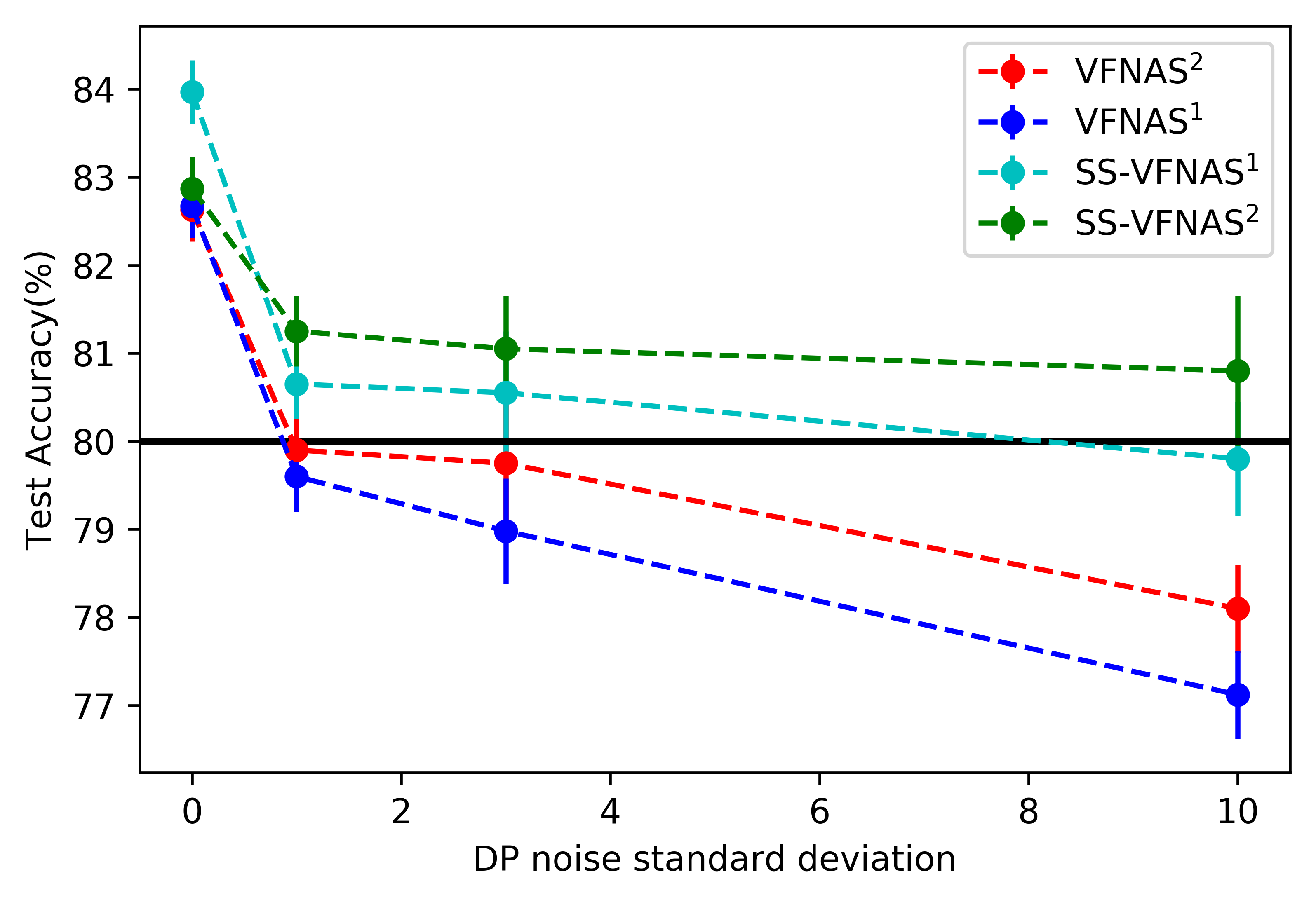}}
\caption{Test Accuracy of different algorithms with different noise variances. The black line represents the baseline test accuracy of SSNAS on the data of party 6.}
\label{dp_comparison}
\end{figure}

In this section, the influence of adding differential privacy to the transmitted messages on the model performance is studied. The experiment is conducted  on FedModelNet40  with  VFNAS$^{1}$-M, VFNAS$^{2}$-M, SS-VFNAS$^{1}$-M and SS-VFNAS$^{2}$-M algorithms. A Gaussian differential private mechanism is employed both in the searching and the evaluation processes and the noise variance ranges are chosen from \{0,1,3,10\}. We compare the test accuracy obtained by different algorithms in Fig. \ref{dp_comparison}. Note that we also present the result of the VFNAS and SSNAS, which is conducted on the party having the labels. Fig. \ref{dp_comparison} shows the test accuracy obtained by these algorithms with different noise variances.  

In practice, the larger the variance of differential privacy noise, the stronger the privacy level. However, large noise variance may greatly degrade the model performance of VFL framework. As shown in the figure, when noise variance reaches 10, the model performance degrades to the point that it is comparable to the SSNAS baseline, and the benefits of VFL vanish. Another key observation from Figure \ref{dp_comparison} is that SS-VFNAS achieves much higher accuracy at the same privacy budget than VFNAS, and the privacy savings amplify as the noise level increases, thanks to the communication savings of SS-VFNAS.

\section{Conclusions and Future Work}\label{sec:future}
In this work, we propose a self-supervised vertical federated NAS method, SS-VFNAS, for multi-domain image classification problems with data privacy preservation. With SS-VFNAS, model designers from different parties can simultaneously optimize their corresponding network architectures without sacrificing data privacy. SS-VFNAS can incorporate different NAS algorithms. We demonstrate the superior performance, efficiency, transferability and scalability of the proposed SS-VFNAS under various privacy levels. Future work includes further improvement of communication efficiency by adopting accelerated convergence methods such as proximal loss \cite{Liu2019ACE}, and co-optimization of model performance and local computation resources.

\section*{Acknowledgments}
This work was partially supported by the National Key Research and Development Program of China under Grant No. 2018AAA0101100. 

\bibliographystyle{IEEEtran}
\bibliography{hfrl}

\begin{thebibliography}{10}
\providecommand{\url}[1]{#1}
\csname url@samestyle\endcsname
\providecommand{\newblock}{\relax}
\providecommand{\bibinfo}[2]{#2}
\providecommand{\BIBentrySTDinterwordspacing}{\spaceskip=0pt\relax}
\providecommand{\BIBentryALTinterwordstretchfactor}{4}
\providecommand{\BIBentryALTinterwordspacing}{\spaceskip=\fontdimen2\font plus
\BIBentryALTinterwordstretchfactor\fontdimen3\font minus
  \fontdimen4\font\relax}
\providecommand{\BIBforeignlanguage}[2]{{%
\expandafter\ifx\csname l@#1\endcsname\relax
\typeout{** WARNING: IEEEtran.bst: No hyphenation pattern has been}%
\typeout{** loaded for the language `#1'. Using the pattern for}%
\typeout{** the default language instead.}%
\else
\language=\csname l@#1\endcsname
\fi
#2}}
\providecommand{\BIBdecl}{\relax}
\BIBdecl

\bibitem{mcmahan2017communication}
B.~McMahan, E.~Moore, D.~Ramage, S.~Hampson, and B.~A. y~Arcas,
  ``Communication-efficient learning of deep networks from decentralized
  data,'' in \emph{Artificial Intelligence and Statistics}.\hskip 1em plus
  0.5em minus 0.4em\relax PMLR, 2017, pp. 1273--1282.

\bibitem{yang2019federated}
Q.~Yang, Y.~Liu, T.~Chen, and Y.~Tong, ``Federated machine learning: Concept
  and applications,'' \emph{ACM Transactions on Intelligent Systems and
  Technology (TIST)}, vol.~10, no.~2, pp. 1--19, 2019.

\bibitem{kairouz2019federated}
B.~A. Peter~Kairouz, H. Brendan~McMahan, ``Advances and open problems in
  federated learning,'' \emph{arXiv preprint arXiv:1912.04977}, 2016.

\bibitem{tensorflowfederated}
T.~T. Authors, ``Tensorflow federated,'' 2019,
  https://www.tensorflow.org/federated.

\bibitem{ryffel2018generic}
T.~Ryffel, A.~Trask, M.~Dahl, B.~Wagner, J.~Mancuso, D.~Rueckert, and
  J.~Passerat-Palmbach, ``A generic framework for privacy preserving deep
  learning,'' \emph{arXiv preprint arXiv:1811.04017}, 2018.

\bibitem{paddlefl}
T.~P. Authors, ``Paddlefl,'' 2019, https://github.com/PaddlePaddle/PaddleFL.

\bibitem{webankfate}
T.~F. Authors, ``Fate,'' 2019, https://github.com/FederatedAI/FATE.

\bibitem{rothchild2020fetchsgd}
D.~Rothchild, A.~Panda, E.~Ullah, N.~Ivkin, I.~Stoica, V.~Braverman,
  J.~Gonzalez, and R.~Arora, ``Fetchsgd: Communication-efficient federated
  learning with sketching,'' \emph{arXiv preprint arXiv:2007.07682}, 2020.

\bibitem{Kairouz2019AdvancesAO}
P.~Kairouz, H.~McMahan, B.~Avent, A.~Bellet, M.~Bennis, A.~Bhagoji,
  K.~Bonawitz, Z.~Charles, G.~Cormode, R.~Cummings, R.~G.~L. D’Oliveira,
  S.~E. Rouayheb, D.~Evans, J.~Gardner, Z.~A. Garrett, A.~Gasc{\'o}n, B.~Ghazi,
  P.~B. Gibbons, M.~Gruteser, Z.~Harchaoui, C.~He, L.~He, Z.~Huo,
  B.~Hutchinson, J.~Hsu, M.~Jaggi, T.~Javidi, G.~Joshi, M.~Khodak,
  J.~Konecn{\'y}, A.~Korolova, F.~Koushanfar, O.~Koyejo, T.~Lepoint, Y.~Liu,
  P.~Mittal, M.~Mohri, R.~Nock, A.~{\"O}zg{\"u}r, R.~Pagh, M.~Raykova, H.~Qi,
  D.~Ramage, R.~Raskar, D.~Song, W.~Song, S.~Stich, Z.~Sun, A.~T. Suresh,
  F.~Tram{\`e}r, P.~Vepakomma, J.~Wang, L.~Xiong, Z.~Xu, Q.~Yang, F.~Yu, H.~Yu,
  and S.~Zhao, ``Advances and open problems in federated learning,''
  \emph{ArXiv}, vol. abs/1912.04977, 2019.

\bibitem{Vepakomma2018Split}
P.~Vepakomma, O.~Gupta, T.~Swedish, and R.~Raskar, ``Split learning for health:
  Distributed deep learning without sharing raw patient data,'' \emph{arXiv
  preprint arXiv:1812.00564}, 2018.

\bibitem{gupta2018distributed}
O.~Gupta and R.~Raskar, ``Distributed learning of deep neural network over
  multiple agents,'' \emph{Journal of Network and Computer Applications}, vol.
  116, pp. 1--8, 2018.

\bibitem{Hu2019LearningPO}
Y.~Hu, P.~Liu, L.~Kong, and D.~Niu, ``Learning privately over distributed
  features: An admm sharing approach,'' \emph{ArXiv}, vol. abs/1907.07735,
  2019.

\bibitem{Chen2020FedHealthAF}
Y.~Chen, J.~Wang, C.~Yu, W.~Gao, and X.~Qin, ``Fedhealth: A federated transfer
  learning framework for wearable healthcare,'' \emph{IEEE Intelligent
  Systems}, vol.~35, pp. 83--93, 2020.

\bibitem{Liu2020ASF}
Y.~Liu, Y.~Kang, C.~Xing, T.~Chen, and Q.~Yang, ``A secure federated transfer
  learning framework,'' \emph{IEEE Intelligent Systems}, vol.~35, pp. 70--82,
  2020.

\bibitem{ebrahimighahnavieh2020deep}
M.~A. Ebrahimighahnavieh, S.~Luo, and R.~Chiong, ``Deep learning to detect
  alzheimer's disease from neuroimaging: A systematic literature review,''
  \emph{Computer Methods and Programs in Biomedicine}, vol. 187, p. 105242,
  2020.

\bibitem{Changqing2018Multi}
Changqing, Zhang, Ehsan, Adeli, Tao, Zhou, Xiaobo, Chen, Dinggang, and Shen,
  ``Multi-layer multi-view classification for alzheimer's disease diagnosis.''
  \emph{Proceedings of the .aaai Conference on Artificial Intelligence.aaai
  Conference on Artificial Intelligence}, 2018.

\bibitem{Liu2019ACE}
Y.~Liu, Y.~Kang, X.~wei Zhang, L.~Li, Y.~Cheng, T.~Chen, M.~Hong, and Q.~Yang,
  ``A communication efficient collaborative learning framework for distributed
  features,'' \emph{arXiv: Learning}, 2019.

\bibitem{wu2020mitigatingba}
C.~Wu, X.~Yang, S.~Zhu, and P.~Mitra, ``Mitigating backdoor attacks in
  federated learning,'' \emph{ArXiv}, vol. abs/2011.01767, 2020.

\bibitem{wei2020aff}
W.~Wei, L.~Liu, M.~Loper, K.-H. Chow, M.~Gursoy, S.~Truex, and Y.~Wu, ``A
  framework for evaluating gradient leakage attacks in federated learning,''
  \emph{ArXiv}, vol. abs/2004.10397, 2020.

\bibitem{elsken2019neural}
T.~Elsken, J.~H. Metzen, and F.~Hutter, ``Neural architecture search: A
  survey,'' \emph{Journal of Machine Learning Research}, vol.~20, no.~55, pp.
  1--21, 2019.

\bibitem{zhu2020real-time}
H.~Zhu and Y.~Jin, ``Real-time federated evolutionary neural architecture
  search.'' \emph{arXiv: Learning}, 2020.

\bibitem{singh2020differentially}
I.~Singh, H.~Zhou, K.~Yang, M.~Ding, B.~Lin, and P.~Xie,
  ``Differentially-private federated neural architecture search,'' \emph{arXiv
  preprint arXiv:2006.10559}, 2020.

\bibitem{xu2020neural}
M.~Xu, Y.~Zhao, K.~Bian, G.~Huang, Q.~Mei, and X.~Liu, ``Neural architecture
  search over decentralized data,'' \emph{arXiv: Learning}, 2020.

\bibitem{chaoyang2020fednas}
S.~A. Chaoyang~He, Murali~Annavaram, ``Fednas: Federated deep learning via
  neural architecture search,'' \emph{arXiv preprint arXiv:2004.08546}, 2020.

\bibitem{Lu2020CommunicationefficientFL}
Y.~Lu, X.~hong Huang, K.~Zhang, S.~Maharjan, and Y.~Zhang,
  ``Communication-efficient federated learning and permissioned blockchain for
  digital twin edge networks,'' \emph{IEEE Internet of Things Journal}, pp.
  1--1, 2020.

\bibitem{Asad2020EvaluatingTC}
M.~Asad, A.~Moustafa, T.~Ito, and A.~Muhammad, ``Evaluating the communication
  efficiency in federated learning algorithms,'' \emph{ArXiv}, vol.
  abs/2004.02738, 2020.

\bibitem{zhu2019deeplf}
L.~Zhu, Z.~Liu, and S.~Han, ``Deep leakage from gradients,'' \emph{ArXiv}, vol.
  abs/1906.08935, 2019.

\bibitem{Hard2018federated}
A.~Hard, K.~Rao, R.~Mathews, S.~Ramaswamy, F.~Beaufays, S.~Augenstein,
  H.~Eichner, C.~Kiddon, and D.~Ramage, ``Federated learning for mobile
  keyboard prediction,'' \emph{arXiv preprint arXiv:1811.03604}, 2018.

\bibitem{Ramaswamy2019federated}
S.~Ramaswamy, R.~Mathews, K.~Rao, and F.~Beaufays, ``Federated learning for
  emoji prediction in a mobile keyboard,'' \emph{arXiv}, 2019.

\bibitem{Chen2019federated}
M.~Chen, R.~Mathews, T.~Ouyang, and F.~Beaufays, ``Federated learning of
  out-of-vocabulary words,'' \emph{arXiv preprint arXiv:1903.10635}, 2019.

\bibitem{Liu2020AsymmetricalVF}
Y.~Liu, X.~Zhang, and L.~Wang, ``Asymmetrical vertical federated learning,''
  \emph{ArXiv}, vol. abs/2004.07427, 2020.

\bibitem{Feng2020MultiParticipantMV}
S.~Feng and H.~Yu, ``Multi-participant multi-class vertical federated
  learning,'' \emph{ArXiv}, vol. abs/2001.11154, 2020.

\bibitem{Yang2019ParallelDL}
S.~Yang, B.~Ren, X.~Zhou, and L.~Liu, ``Parallel distributed logistic
  regression for vertical federated learning without third-party coordinator,''
  \emph{ArXiv}, vol. abs/1911.09824, 2019.

\bibitem{vepakomma2019reducing}
P.~Vepakomma, O.~Gupta, A.~Dubey, and R.~Raskar, ``Reducing leakage in
  distributed deep learning for sensitive health data,'' \emph{arXiv preprint
  arXiv:1812.00564}, 2019.

\bibitem{poirot2019split}
M.~G. Poirot, P.~Vepakomma, K.~Chang, J.~Kalpathy-Cramer, R.~Gupta, and
  R.~Raskar, ``Split learning for collaborative deep learning in healthcare,''
  \emph{arXiv preprint arXiv:1912.12115}, 2019.

\bibitem{Ceballos2020SplitNNdrivenVP}
I.~Ceballos, V.~Sharma, E.~M{\'u}gica, A.~Singh, A.~Rom{\'a}n, P.~Vepakomma,
  and R.~Raskar, ``Splitnn-driven vertical partitioning,'' \emph{ArXiv}, vol.
  abs/2008.04137, 2020.

\bibitem{Abuadbba2020CanWU}
S.~Abuadbba, K.~yeon Kim, M.~Kim, C.~Thapa, S.~Çamtepe, Y.~Gao, H.~Kim, and
  S.~Nepal, ``Can we use split learning on 1d cnn models for privacy preserving
  training?'' \emph{ArXiv}, vol. abs/2003.12365, 2020.

\bibitem{liu2018progressive}
C.~Liu, B.~Zoph, M.~Neumann, J.~Shlens, W.~Hua, L.-J. Li, L.~Fei-Fei,
  A.~Yuille, J.~Huang, and K.~Murphy, ``Progressive neural architecture
  search,'' in \emph{Proceedings of the European Conference on Computer Vision
  (ECCV)}, 2018, pp. 19--34.

\bibitem{zoph2018learning}
B.~Zoph, V.~Vasudevan, J.~Shlens, and Q.~V. Le, ``Learning transferable
  architectures for scalable image recognition,'' in \emph{Proceedings of the
  IEEE conference on computer vision and pattern recognition}, 2018, pp.
  8697--8710.

\bibitem{real2018regularized}
E.~Real, A.~Aggarwal, Y.~Huang, and Q.~V. Le, ``Regularized evolution for image
  classifier architecture search,'' \emph{arXiv: Neural and Evolutionary
  Computing}, 2018.

\bibitem{liu2018darts:}
H.~Liu, K.~Simonyan, and Y.~Yang, ``Darts: Differentiable architecture
  search,'' \emph{arXiv: Learning}, 2018.

\bibitem{Liang2019EvolutionaryNA}
J.~Liang, E.~Meyerson, B.~Hodjat, D.~Fink, K.~Mutch, and R.~Miikkulainen,
  ``Evolutionary neural automl for deep learning,'' \emph{Proceedings of the
  Genetic and Evolutionary Computation Conference}, 2019.

\bibitem{cai2020once}
\BIBentryALTinterwordspacing
H.~Cai, C.~Gan, T.~Wang, Z.~Zhang, and S.~Han, ``Once for all: Train one
  network and specialize it for efficient deployment,'' in \emph{International
  Conference on Learning Representations}, 2020. [Online]. Available:
  \url{https://arxiv.org/pdf/1908.09791.pdf}
\BIBentrySTDinterwordspacing

\bibitem{Cai2019ProxylessNASDN}
H.~Cai, L.~Zhu, and S.~Han, ``Proxylessnas: Direct neural architecture search
  on target task and hardware,'' \emph{ArXiv}, vol. abs/1812.00332, 2019.

\bibitem{Deb2002fast}
K.~{Deb}, A.~{Pratap}, S.~{Agarwal}, and T.~{Meyarivan}, ``A fast and elitist
  multiobjective genetic algorithm: Nsga-ii,'' \emph{IEEE Transactions on
  Evolutionary Computation}, vol.~6, no.~2, pp. 182--197, 2002.

\bibitem{krizhevsky2009learning}
A.~Krizhevsky, G.~Hinton \emph{et~al.}, ``Learning multiple layers of features
  from tiny images,'' 2009.

\bibitem{russakovsky2015imagenetls}
O.~Russakovsky, J.~Deng, H.~Su, J.~Krause, S.~Satheesh, S.~Ma, Z.~Huang,
  A.~Karpathy, A.~Khosla, M.~S. Bernstein, A.~Berg, and L.~Fei-Fei, ``Imagenet
  large scale visual recognition challenge,'' \emph{International Journal of
  Computer Vision}, vol. 115, pp. 211--252, 2015.

\bibitem{Liu2020AreLN}
C.~Liu, P.~Doll{\'a}r, K.~He, R.~B. Girshick, A.~Yuille, and S.~Xie, ``Are
  labels necessary for neural architecture search?'' \emph{ArXiv}, vol.
  abs/2003.12056, 2020.

\bibitem{Kaplan2020SelfsupervisedNA}
S.~Kaplan and R.~Giryes, ``Self-supervised neural architecture search,''
  \emph{ArXiv}, vol. abs/2007.01500, 2020.

\bibitem{He2020MiLeNASEN}
C.~He, H.~Ye, L.~Shen, and T.~Zhang, ``Milenas: Efficient neural architecture
  search via mixed-level reformulation,'' \emph{2020 IEEE/CVF Conference on
  Computer Vision and Pattern Recognition (CVPR)}, pp. 11\,990--11\,999, 2020.

\bibitem{dwork2006calibrating}
C.~Dwork, F.~McSherry, K.~Nissim, and A.~Smith, ``Calibrating noise to
  sensitivity in private data analysis,'' in \emph{Theory of cryptography
  conference}.\hskip 1em plus 0.5em minus 0.4em\relax Springer, 2006, pp.
  265--284.

\bibitem{dwork2011firm}
C.~Dwork, ``A firm foundation for private data analysis,'' \emph{Communications
  of the ACM}, vol.~54, no.~1, pp. 86--95, 2011.

\bibitem{dwork2014algorithmic}
C.~Dwork, A.~Roth \emph{et~al.}, ``The algorithmic foundations of differential
  privacy.'' \emph{Foundations and Trends in Theoretical Computer Science},
  vol.~9, no. 3-4, pp. 211--407, 2014.

\bibitem{bhowmick2018protection}
A.~Bhowmick, J.~Duchi, J.~Freudiger, G.~Kapoor, and R.~Rogers, ``Protection
  against reconstruction and its applications in private federated learning,''
  \emph{arXiv preprint arXiv:1812.00984}, 2018.

\bibitem{zhu2019deep}
L.~Zhu, Z.~Liu, and S.~Han, ``Deep leakage from gradients,'' in \emph{Advances
  in Neural Information Processing Systems}, 2019, pp. 14\,774--14\,784.

\bibitem{abadi2016deep}
M.~Abadi, A.~Chu, I.~Goodfellow, H.~B. McMahan, I.~Mironov, K.~Talwar, and
  L.~Zhang, ``Deep learning with differential privacy,'' in \emph{Proceedings
  of the 2016 ACM SIGSAC Conference on Computer and Communications Security},
  2016, pp. 308--318.

\bibitem{Chen2020ASF}
T.~Chen, S.~Kornblith, M.~Norouzi, and G.~E. Hinton, ``A simple framework for
  contrastive learning of visual representations,'' \emph{ArXiv}, vol.
  abs/2002.05709, 2020.

\bibitem{chen2020improvedbw}
X.~Chen, H.~Fan, R.~B. Girshick, and K.~He, ``Improved baselines with momentum
  contrastive learning,'' \emph{ArXiv}, vol. abs/2003.04297, 2020.

\bibitem{he2020Momentumcf}
K.~He, H.~Fan, Y.~Wu, S.~Xie, and R.~B. Girshick, ``Momentum contrast for
  unsupervised visual representation learning,'' \emph{2020 IEEE/CVF Conference
  on Computer Vision and Pattern Recognition (CVPR)}, pp. 9726--9735, 2020.

\bibitem{Oord2018RepresentationLW}
A.~Oord, Y.~Li, and O.~Vinyals, ``Representation learning with contrastive
  predictive coding,'' \emph{ArXiv}, vol. abs/1807.03748, 2018.

\bibitem{he2016deep}
K.~He, X.~Zhang, S.~Ren, and J.~Sun, ``Deep residual learning for image
  recognition,'' in \emph{Proceedings of the IEEE conference on computer vision
  and pattern recognition}, 2016, pp. 770--778.

\bibitem{iandola2016squeezenet}
F.~N. Iandola, S.~Han, M.~W. Moskewicz, K.~Ashraf, W.~J. Dally, and K.~Keutzer,
  ``Squeezenet: Alexnet-level accuracy with 50x fewer parameters and< 0.5 mb
  model size,'' \emph{arXiv preprint arXiv:1602.07360}, 2016.

\bibitem{zhang2018shufflenet}
X.~Zhang, X.~Zhou, M.~Lin, and J.~Sun, ``Shufflenet: An extremely efficient
  convolutional neural network for mobile devices,'' in \emph{Proceedings of
  the IEEE conference on computer vision and pattern recognition}, 2018, pp.
  6848--6856.

\bibitem{Wu20153D}
Z.~Wu, S.~Song, A.~Khosla, F.~Yu, L.~Zhang, X.~Tang, and J.~Xiao, ``3d
  shapenets: A deep representation for volumetric shapes,'' in
  \emph{Proceedings of the IEEE conference on computer vision and pattern
  recognition}, 2015, pp. 1912--1920.

\bibitem{Irvin2019CheXpertAL}
J.~Irvin, P.~Rajpurkar, M.~Ko, Y.~Yu, S.~Ciurea-Ilcus, C.~Chute, H.~Marklund,
  B.~Haghgoo, R.~Ball, K.~Shpanskaya, J.~Seekins, D.~Mong, S.~Halabi,
  J.~Sandberg, R.~Jones, D.~Larson, C.~Langlotz, B.~N. Patel, M.~Lungren, and
  A.~Ng, ``Chexpert: A large chest radiograph dataset with uncertainty labels
  and expert comparison,'' in \emph{AAAI}, 2019.

\bibitem{Su2015Multi}
H.~Su, S.~Maji, E.~Kalogerakis, and E.~Learned-Miller, ``Multi-view
  convolutional neural networks for 3d shape recognition,'' in
  \emph{Proceedings of the IEEE international conference on computer vision},
  2015, pp. 945--953.

\end{thebibliography}
\ifCLASSOPTIONcaptionsoff
  \newpage
\fi

\end{document}